\documentclass{article}

\usepackage{arxiv}
\usepackage{amsmath} 
\usepackage[utf8]{inputenc} 
\usepackage[T1]{fontenc}    
\usepackage{hyperref}       
\hypersetup{
    colorlinks=true,   
    linkcolor=blue,
    urlcolor=blue,
    citecolor=blue
}

\usepackage{url}            
\usepackage{booktabs}       
\usepackage{amsfonts}       
\usepackage{nicefrac}       
\usepackage{microtype}      
\usepackage{lipsum}
\usepackage{graphicx}
\graphicspath{ {./images/} }

\usepackage{multirow}
\usepackage{array}
\usepackage{caption}
\usepackage{pifont} 
\usepackage{placeins}
\usepackage[most]{tcolorbox}
\usepackage[table]{xcolor}
\definecolor{lightblue}{RGB}{230,240,255} 
\usepackage{makecell} 
\usepackage{dblfloatfix}

\definecolor{checkgreen}{RGB}{50,150,50} 
\definecolor{crossred}{RGB}{220,80,80}  
\definecolor{mixedblue}{RGB}{100,100,220}

\newcommand{\cmark}{\textcolor{checkgreen}{\ding{51}}}
\newcommand{\xmark}{\textcolor{crossred}{\ding{55}}}
\newcommand{\tmark}{\textcolor{mixedblue}{\ding{51}\kern-0.63em\ding{55}}}

\usepackage{xcolor}
\usepackage{fontawesome5}

\title{H2HMem: A Multimodal Memory Benchmark for Agents in Human--Human Interactions}

\author{
  Shiping Zhu \\
  Jilin University \\
  \texttt{zhusp9923@mails.jlu.edu.cn} \\
  \And
  Yibo Yang$^{\dagger}$ \\
  Shanghai Jiao Tong University\\
  \texttt{yibo.yang93@gmail.com} \\
  \And
  Zhengyang Wang \\
  Jilin University \\
  \texttt{zhengyangw9923@mails.jlu.edu.cn} \\
  \AND
  Tiancheng Shen \\
  University of California at Merced \\
  \texttt{stc199506@gmail.com} \\
  \And
  Dandan Guo$^{\dagger}$  \\
  Jilin University \\
  \texttt{gdd\_xidian@126.com} \\
  \And
  Ming-Hsuan Yang \\
  University of California at Merced \\
  \texttt{minghsuanyang@gmail.com} \\
}

\begin{document}
\maketitle

\renewcommand{\thefootnote}{}
\footnotetext{$^\dagger$denotes the corresponding author.}

\newtcbox{\roundbutton}[1][]{
  on line,
  tcbox raise base,
  arc=7pt,
  boxrule=0.5pt,
  left=6pt,
  right=6pt,
  top=2pt,
  bottom=2pt,
  nobeforeafter,
  #1
}

\newcommand{\buttonlink}[3]{%
\begingroup
\hypersetup{pdfborder={0 0 0}}
\href{#2}{%
\roundbutton[#1]{#3}%
}
\endgroup
}
\vspace{-0.5cm}
\begin{center}
\buttonlink{colback=blue!5,colframe=blue!35}
{https://github.com/varib1/H2HMEM}
{\textcolor{blue!80!black}{\faGithub\ \textbf{Code}}}
\hspace{0.35cm}
\buttonlink{colback=orange!8,colframe=orange!45}
{https://huggingface.co/datasets/varib/H2HMEM}
{\textcolor{orange!90!black}{\faDatabase\ \textbf{Dataset}}}
\hspace{0.35cm}
\buttonlink{colback=green!8,colframe=green!45}
{https://h2hmemleaderboard1.vercel.app/}
{\textcolor{green!50!black}{\faTrophy\ \textbf{Leaderboard}}}
\hspace{0.35cm}
\buttonlink{colback=purple!8,colframe=purple!45}
{https://h2hmemprojectpage.vercel.app/}
{\textcolor{purple!80!black}{\faFile* \ \textbf{Project Page}}}
\end{center}
\vspace{0.15cm}

\begin{abstract}
Large language model agents are increasingly deployed in human--human interaction settings, such as meeting assistants and clinical documentation systems, where they must observe conversations and retain information for downstream queries. Unlike traditional human--assistant settings, these environments are inherently multimodal, involve complex discourse phenomena such as anaphora and deixis, and contain asynchronous or conflicting information from multiple participants. However, existing memory benchmarks largely focus on single-user, text-only interactions, failing to capture these challenges. To address this gap, we introduce \textbf{H2HMem}, a \textbf{Human-to-Human Multimodal Memory Benchmark} for evaluating memory capabilities in complex human--human interactions. H2HMem includes both dyadic and multi-party conversations with multimodal information streams, and evaluates agents along three dimensions: memory recall, reasoning, and application. Experiments with advanced agents reveal substantial limitations in constructing, retaining, and utilizing memories across modalities, participants, and sessions, highlighting substantial room for improvement in next-generation LLM agents.
\end{abstract}
\section{Introduction}

\begin{figure}[htbp]
\centering
\includegraphics[width=\textwidth]{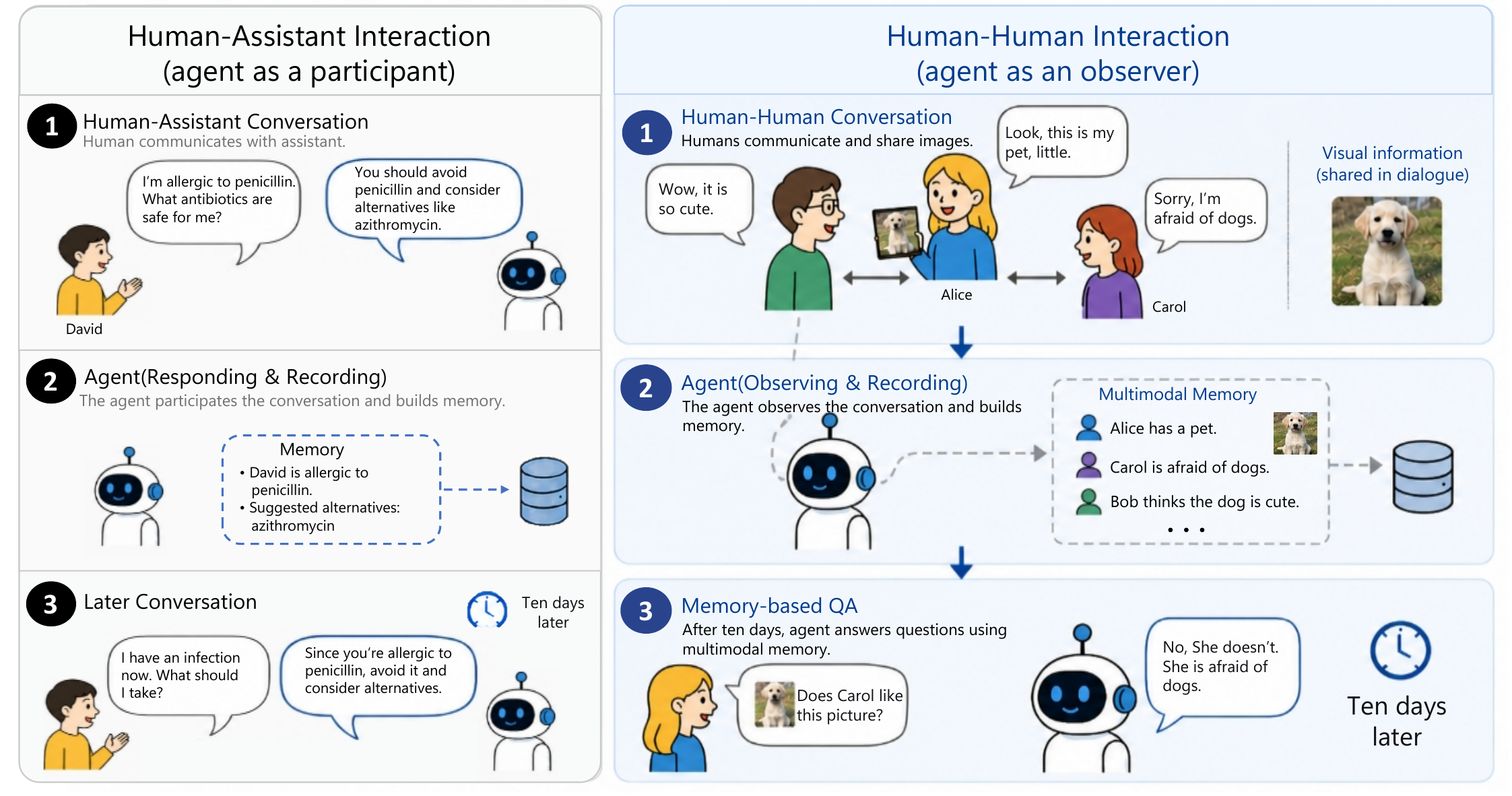}
\caption{Comparison between Human--Assistant Interaction and Human--Human Interaction.}
\label{fig:setting_comparison}
\end{figure}

Large language model (LLM) agents, such as ChatGPT~\cite{openai2023chatgpt} and DeepSeek~\cite{deepseek2024v3}, have advanced substantially, with memory mechanisms improving coherence in extended multi-turn human--assistant dialogues~\cite{du2026memoryautonomousllmagentsmechanisms}. Beyond this paradigm, however, a new class of applications is emerging, as illustrated in Figure~\ref{fig:setting_comparison}: \textbf{LLM agents as observers in human--human interactions}. In these settings, agents passively capture critical conversational information for subsequent querying. This capability underpins growing real-world applications, including clinical documentation systems that generate patient-centered notes from clinician--patient dialogues~\cite{Biswas_2024, vaid2026bertaopensourcemodulartool, CDT148228}, AI-powered medical board meeting assistants processing multimodal inputs~\cite{11346322}, and general meeting summarization systems~\cite{Asthana_2025}. To operate effectively, such agents must track information distributed across multiple participants, maintain context over extended interactions, and integrate signals across modalities~\cite{zhu2025overhearingllmagentssurvey}. Robust multimodal memory is therefore essential.

These emerging deployment environments introduce three fundamental challenges. First, human–human conversations are inherently multimodal, naturally interleaving text with visual content such as shared photographs and screen captures~\cite{lee-etal-2024-dialogcc, Guo2024AnEM}. Second, natural language exhibits complex phenomena—such as anaphora and discourse deixis—that require agents to resolve references against an evolving conversational memory rather than retrieve isolated facts~\cite{Kim2022PipelineCR}. Third, these interactions often involve multiple participants (dyadic or multi-party) who jointly shape the dialogue, contributing information asynchronously and at times presenting conflicting perspectives~\cite{abbo2025fastmultipartyopenendedconversation, liu2025socialgazellmsliterature}. Systematically evaluating memory mechanisms under these conditions is therefore essential.

Existing memory benchmarks, however, fail to capture these complexities. Most are designed for single-user, text-only human–assistant interactions~\cite{wu2025longmemeval, tan-etal-2025-membench, hu2026evaluating, shen2026evolmemcognitivedrivenbenchmarkmultisession}. Although recent efforts have begun exploring human–human conversations, they remain limited in scope: LoCoMo~\cite{Maharana2024EvaluatingVL} incorporates vision but is restricted to dyadic interactions and lacks a comprehensive memory evaluation framework, whereas others~\cite{hu2026evaluatinglonghorizonmemorymultiparty} support multi-party settings but remain exclusively text-based. Consequently, no existing benchmark adequately captures the full spectrum of human–human interactions—spanning both dyadic and multi-party settings—while enabling multimodal memory evaluation. A comparison between existing benchmarks and ours is presented in Table~\ref{tab:benchmark_comparison}.

To address this gap, we introduce \textbf{H2HMem}, a \textbf{Human-to-Human Multimodal Memory Benchmark}. Since directly collecting real-world multimodal conversations raises substantial privacy concerns that are difficult to fully mitigate through de-identification~\cite{hakobyan2025development, ietf-vcon-overview-01}, we develop a human-in-the-loop generation pipeline. By guiding LLM agents to iteratively generate dialogues, this pipeline avoids privacy risks while producing realistic multimodal, multi-session, and multi-participant interactions. We design evaluation tasks along three functional dimensions of memory that reflect the complexities of natural communication: (1) \textbf{Memory Recall}, which measures retrieval of multimodal facts and resolution of evolving knowledge across sessions, including Unimodal Precise Recall (UPR), Cross-modal Related Retrieval (CRR), and Knowledge Resolution (KR); (2) \textbf{Memory Reasoning}, which evaluates higher-level inference through Multimodal Causal Reasoning (MCR), Reference \& Evolution Tracking (RET), and Temporal Reasoning (TR); and (3) \textbf{Memory Application}, which assesses the ability to use memory in dynamic settings through Test-Time Learning (TTL), Conflict Detection (CD), and Answer Refusal (AR). Together, these dimensions provide a comprehensive framework that moves beyond simple recall to systematically evaluate memory in complex human--human interactions. We summarize our contributions as follows:
\begin{itemize}
\item We introduce H2HMem, a benchmark for evaluating multimodal memory in realistic human--human observer scenarios, covering both dyadic and multi-party interactions.
\item We construct a large-scale multimodal, multi-session dataset through a privacy-preserving human-in-the-loop pipeline that captures the evolving nature of real-world communication.
\item We propose a comprehensive evaluation taxonomy spanning recall, reasoning, and application, revealing key limitations of current MLLMs in cross-modal memory alignment and structured reasoning.
\end{itemize}

\begin{table}[htbp]
\centering
\small
\setlength{\tabcolsep}{2pt}  
\renewcommand{\arraystretch}{1.2}  
\caption{Comparison of dialogue benchmarks. \cmark: fully covered; \xmark: not covered; \tmark: partially covered. A. Round and A. Img. denote the average number of rounds and images per session, respectively, and MM-Info. indicates whether multimodal information is included.}

\begin{tabular}{@{} l|c|ccc|ccc|ccc|ccc @{}}
\toprule
\multirow{2}{*}{Benchmark} 
& \multirow{2}{*}{Interaction Type} 
& \multicolumn{3}{c|}{Conversational Characteristics} 
& \multicolumn{3}{c|}{Recall} 
& \multicolumn{3}{c|}{Reasoning} 
& \multicolumn{3}{c}{Application} \\
\cmidrule(lr){3-5} \cmidrule(lr){6-8} \cmidrule(lr){9-11} \cmidrule(lr){12-14}
& 
& A. Round & A. Img. & MM-Info. 
& UPR & CRR & KR 
& MCR & RET & TR 
& TTL & CD & AR \\
\midrule
LongMemEval~\cite{wu2025longmemeval}       & Human--assistant   & 5.19  & --   & \xmark & \tmark & \xmark & \cmark & \tmark & \xmark & \cmark & \xmark & \xmark & \cmark \\
PersonaMem~\cite{jiang2025knowmerespondme}        & Human--assistant   & 15--30& --   & \xmark & \tmark & \xmark & \cmark & \xmark & \xmark & \cmark & \xmark & \xmark & \xmark \\
Mem-Gallery~\cite{bei2026memgallerybenchmarkingmultimodallongterm}       & Human--assistant   & 16.51 & 4.18 & \cmark & \cmark & \tmark & \cmark & \xmark & \xmark & \cmark & \cmark & \cmark & \cmark \\
MemoryAgentBench~\cite{hu2026evaluating}  & Human--assistant   & 9.55  & --   & \xmark & \tmark & \xmark & \xmark & \xmark & \xmark & \tmark & \cmark & \cmark & \xmark \\
LoCoMo~\cite{Maharana2024EvaluatingVL}            & Dyadic           & 10.81 & 3.35 & \cmark & \cmark & \xmark & \xmark & \xmark & \xmark & \cmark & \xmark & \xmark & \cmark \\
MSC~\cite{xu-etal-2022-beyond}               & Dyadic           & 8.16  & --   & \xmark & \tmark & \xmark & \xmark & \xmark & \xmark & \xmark & \xmark & \xmark & \xmark \\
EverMemBench~\cite{hu2026evaluatinglonghorizonmemorymultiparty}      & Multi-party      & 28.0  & --   & \xmark & \tmark & \xmark & \xmark & \tmark & \xmark & \cmark & \xmark & \xmark & \xmark \\
\midrule
\textbf{H2HMem}     & \textbf{Dyadic, Multi-party} & 22.91 & 4.21 & \cmark & \cmark & \cmark & \cmark & \cmark & \cmark & \cmark & \cmark & \cmark & \cmark \\
\bottomrule
\end{tabular}
\label{tab:benchmark_comparison}
\end{table}

\section{Related Work}

\textbf{Agents in human--human Interactions.}
Recent work has begun to study LLM-based agents in human--human interaction settings, 
where the agent acts as an observer over continuous conversational streams~\cite{Asthana_2025}. 
Unlike traditional human--assistant scenarios, these settings require persistent interpretation 
of evolving human--human interactions and maintaining coherence over long temporal horizons~\cite{chen-etal-2025-llamapie, 10.1145/3715336.3735833}. 
As agents are deployed in increasingly rich environments, multimodal inputs—including speech, text, and documents—are incorporated~\cite{11346322}, significantly increasing the complexity of information flow. 
Commercial systems such as \textit{Zoom AI Companion}~\cite{zoom2026} already reflect this trend, 
integrating multimodal meeting content for downstream querying. 
These characteristics jointly impose strong requirements on an agent’s ability to track, integrate, and retain information over time, 
making memory a central capability in such settings.

\noindent \textbf{Memory Mechanisms for LLM Agents.} Existing memory methods for LLM agents mainly fall into three paradigms. 
One line of work extends the context window by directly incorporating long interaction histories into the model input~\cite{Kitaev2020Reformer:, su2024roformer, 10.5555/3600270.3601459}. 
Although simple, this approach incurs high computational cost, suffers from long-context degradation~\cite{chen2023extending}, and lacks cross-session persistence. 
Another line of work adopts retrieval-augmented generation, maintaining an external memory store for history retrieval~\cite{lewis2020retrieval, wang2023augmenting, shi2024replug}. 
While scalable and persistent, such methods mainly support factual recall and struggle with episodic dependencies and causal structures~\cite{rajesh2026beyond}. 
A third direction introduces specialized memory modules with explicit operations such as writing, indexing, summarization, and forgetting~\cite{xu2026amem, kang-etal-2025-memory}. 
Despite these advances, existing methods are primarily developed and evaluated in human--assistant settings, leaving their effectiveness in multimodal human--human interactions unclear.

\noindent \textbf{Memory Benchmarks.}
To evaluate memory capabilities, a range of benchmarks have been proposed. 
In human--assistant settings, \textsc{PersonaMem}~\cite{jiang2025knowmerespondme} studies preference following and user profiling, 
\textsc{LongMemEval}~\cite{wu2025longmemeval} focuses on long-term memory in multi-turn dialogue, 
\textsc{Mem-Gallery}~\cite{bei2026memgallerybenchmarkingmultimodallongterm} extends to multimodal interactions, 
and \textsc{MemoryAgentBench}~\cite{hu2026evaluating} evaluates memory in dialogue streams. 
Moving toward human--human settings, \textsc{MSC}~\cite{xu-etal-2022-beyond} and \textsc{LoCoMo}~\cite{Maharana2024EvaluatingVL} 
consider conversational memory, but both are restricted to dyadic interactions. 
\textsc{EverMemBench}~\cite{hu2026evaluatinglonghorizonmemorymultiparty} extends to multi-party dialogue, 
yet leaves multimodal aspects underexplored. 
Related work has also considered observer-style agents: \textsc{MemBench}~\cite{tan-etal-2025-membench} 
studies passive observation with one-sided inputs, and \textsc{M3-Bench}~\cite{long2025seeinglisteningrememberingreasoning} 
introduces video-based QA over human interactions, but is constrained by limited temporal scope. 
Overall, no existing benchmark jointly captures multimodality, dyadic \& multi-party interaction, and long-horizon memory
in human--human settings within an integrated evaluation framework. Table~\ref{tab:benchmark_comparison} highlights these gaps and situates our proposed H2HMem. 

\section{H2HMem}
\label{headings}
\label{gen_inst}
\begin{figure}[t]
    \centering
    \includegraphics[width=\textwidth]{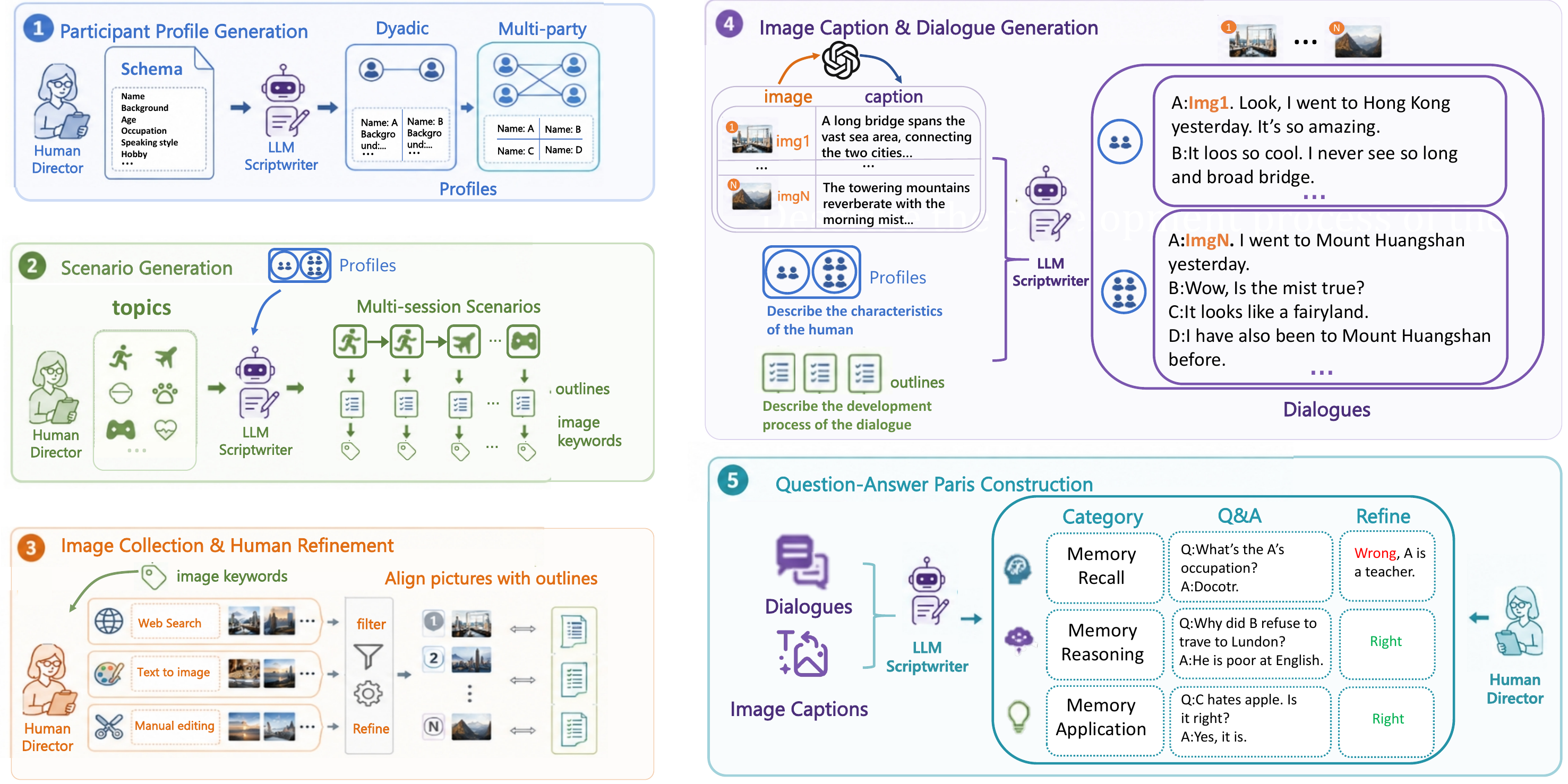}
    \caption{Dataset construction pipeline of our H2HMem. (1) generating dyadic and multi-party participant profiles from structured schemas; (2) creating multi-session scenarios with topic-specific outlines and image keywords; (3) collecting and refining images to align visual evidence with scenario outlines; (4) prompting an LLM scriptwriter with profiles, outlines, and image captions to generate dialogues; and (5) constructing and human-verifying question–answer pairs across memory recall, reasoning, and application. }
    \label{fig:pipeline}
\end{figure}
\subsection{Problem Formulation}

We study multi-session, multimodal question answering grounded in human--human interactions. 
An interaction, denoted as a dialogue $S$, is represented as a sequence of $T$ sessions, 
$S = (s_1, \dots, s_T)$. Each session $s_t$, associated with timestamp $\tau_t$, 
corresponds to a single-day conversation, typically centered around a specific topic 
(e.g., movie, pet, or health). 

A session is defined as $s_t = (u_{t,1}, \dots, u_{t,n_t})$, where $n_t$ is the number of 
chronologically ordered utterances. Each utterance is a multimodal tuple 
$u_{t,i} = (p_{t,i}, x_{t,i}, v_{t,i})$, where $p_{t,i} \in \mathcal{P}$ denotes the speaker, 
$x_{t,i}$ is the textual content, and $v_{t,i}$ is an optional image. 
The participant set $\mathcal{P}$ determines the interaction type: the dialogue is dyadic 
if $|\mathcal{P}| = 2$ and multi-party if $|\mathcal{P}| \geq 3$. 
An utterance is valid if $x_{t,i} \neq \emptyset$ or $v_{t,i} \neq \emptyset$. 

We formalize memory as follows. A storage function maps each utterance $u_{t,i}$ to a memory unit $m$. 
After processing all sessions, the memory state is
\[
\mathcal{M}_T = \{m_1, \dots, m_N\}, \quad N = \sum_{k=1}^{T} n_k.
\]
Given a query $q$, the system retrieves a subset 
$\mathcal{R} = \operatorname{retrieve}(q, \mathcal{M}_T)$ and produces the final answer 
$a = \operatorname{LLM}(\mathcal{R}, q)$. 

\begin{table}[htbp]
\centering
\fontsize{9pt}{9pt}\selectfont
\setlength{\tabcolsep}{6pt}
\renewcommand{\arraystretch}{1.0}
\caption{Statistics of the H2HMem dataset. ``Conv. Data'' denotes conversation data. ``Eval. Data'' denotes evaluation data. ``Avg. sessions/dialogue'' denotes the average number of sessions per dialogue. ``Avg. rounds/session'' denotes the average number of dialogue rounds per session. ``Avg. participants/dialogue'' denotes the average number of participants per dialogue.}
\label{tab:H2HMem}
\begin{tabular}{l l c c c}
\toprule
\textbf{H2HMem} & \textbf{Aspect} & \textbf{Dyadic} & \textbf{Multi-party} & \textbf{Sum} \\
\midrule

\multirow{7}{*}{Conv. Data}
& Dialogues        & 20     & 5     & 25 \\
& Sessions         & 284    & 25    & 309 \\
& Dialogue Rounds  & 5,316  & 1,762 & 7,078 \\
& Included Images  & 951    & 349   & 1,300 \\
\cmidrule(lr){2-5}
& Avg. participants/dialogue & 2.0   & 5.2   & 2.64 \\
& Avg. sessions/dialogue & 14.2   & 5.0   & 12.4 \\
& Avg. rounds/session & 18.7   & 70.5  & 22.9 \\
\midrule

\multirow{2}{*}{Eva. Data}
& QA Pairs         & 2,046  & 190   & 2,236 \\
& Included Images  & 596    & 22    & 618 \\

\bottomrule
\end{tabular}
\end{table}

\subsection{Dataset Construction Pipeline}
We construct the dataset modeling human--human interactions under an online conversational setting, including dyadic and multi-party interactions. The overall statistics of H2HMem benchmark are presented in Table~\ref{tab:H2HMem}. More detailed conversation data statistics is shown in Appendix~\ref{app:sec1.1}. Both dialogue types follow the same pipeline with minor parameter differences. We adopt a human-in-the-loop paradigm: humans act as directors, ensuring scenario consistency, visual grounding, and quality control; LLMs serve as scriptwriters, generating dialogues, scenarios, and QA pairs. An overview of our pipeline is shown in Figure~\ref{fig:pipeline}.
\begin{figure}[htbp]
    \centering
    \includegraphics[width=\textwidth]{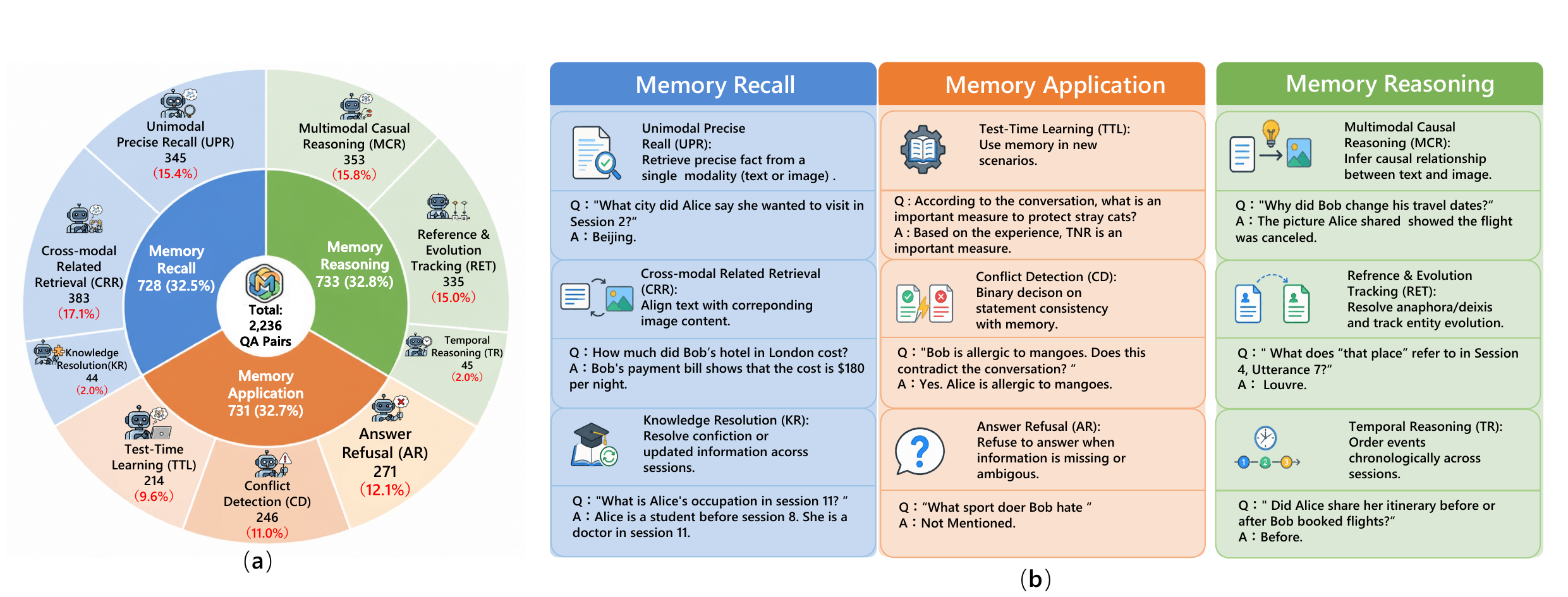}
    \caption{Figure (a) shows the total number and distribution of questions; Figure (b) provides definition and an example for each question type. }
    \label{fig:problem}
\end{figure}

\noindent \textbf{Online Conversational Setting.}
We focus on online conversational environments, where interactions occur via temporally ordered messages, allowing asynchronous participation, as in social media or messaging platforms~\cite{reece2023candor, zheng-etal-2022-mmchat}. This setting offers three key advantages: strong ecological validity, structured information flow, and support for diverse topics and participants yielding richer conversational dynamics~\cite{wang2021naturalconv, lee2026multi}.

\noindent \textbf{Stage 1: Participant Profile Generation.}
We first define a structured schema for participant profiles, inspired by schema-guided dialogue modeling and structured persona-based datasets~\cite{rastogi2020towards, zheng2019personalized}. These profiles include attributes such as personality, background and communication style. Conditioned on this schema, we employ DeepSeek-V3~\cite{deepseek2024v3} to generate structured participant profiles. An example of profiles is presented in Appendix~\ref{app:sec1.2}. 

\noindent \textbf{Stage 2: Scenario Construction.}
We summarize eleven common conversational topics and, given participant profiles, prompt the LLM to sample topics. For each topic, the LLM generates multiple session-level outlines, each describing a session's local events. These sessions are temporally ordered, forming a coherent multi-session scenario $S = (s_1, \dots, s_T)$. The LLM also generates image retrieval keywords to facilitate visual content collection in the subsequent stage. An example of outlines is presented in Appendix~\ref{app:sec1.3}. 

\noindent \textbf{Stage 3: Image Collection and Human Refinement.}
We retrieve images through online search, supplementing retrieval with text-to-image generation~\cite{ramesh2022hierarchicaltextconditionalimagegeneration, saharia2022photorealistictexttoimagediffusionmodels} and manual creation/editing based on the image keywords. Then we filter and refine pictures to align images with outlines, modifying the outlines when necessary. These images become the visual content $v_{t,i}$ in each utterance. More details are presented in Appendix~\ref{app:sec1.4.1}.

\noindent \textbf{Stage 4: Image Captioning and Dialogue Generation.}
Dialogues are generated using DeepSeek-V3~\cite{deepseek2024v3}, conditioned on participant profiles, session outlines, and images. Since DeepSeek-V3~\cite{deepseek2024v3} cannot process images directly, we generate detailed captions via GPT-4o~\cite{openai2024gpt4ocard}. The agent generates dialogues and refers to images using numeric identifiers. We denote each utterance as \(u_{t,i} = (p_{t,i}, x_{t,i}, v_{t,i})\), where \(x_{t,i}\) denotes the textual content and \(v_{t,i}\) denotes the corresponding image obtained by replacing the numeric reference in \(x_{t,i}\) with the actual image.

\noindent \textbf{Stage 5: Question-Answer Pairs Construction.}
Based on the generated dialogues $S = (s_1, \dots, s_T)$, we use DeepSeek-V3~\cite{deepseek2024v3} to generate a diverse set of questions $q$ targeting different memory capabilities (recall, reasoning, application). During generation, any visual information in the dialogues is still replaced with captions. The generated question-answer pairs are further refined by human annotators to ensure clarity, correctness, and appropriate difficulty. More details on the refinement are presented in Appendix~\ref{app:sec1.4.2}. 

\begin{table}[htbp]
\centering
\fontsize{7pt}{9pt}\selectfont
\setlength{\tabcolsep}{6pt}
\renewcommand{\arraystretch}{1.0}
\newcommand{\dmcell}{\cellcolor{lightblue!80}} 
\caption{LLM-Judge performance with GPT-4.1-Nano~\cite{openai2024gpt4technicalreport}. D = Dyadic, M = Multi-party, and D\&M denotes the weighted average, weighted by the number of questions. {\color{blue}\textbf{*}} indicates the higher value between D and M. Bold numbers in the D\&M column denote the best overall performance. Light blue shading highlights D\&M cells. Additional results on other backbone models are reported in Appendix~\ref{app:sec4.1}.}
\label{tab:dyadic_multiparty_combined}
\begin{tabular}{@{} l l l *{10}{l} @{}}
\toprule
\multirow{2}{*}{Category} & \multirow{2}{*}{Method} & \multirow{2}{*}{Dataset} & 
\multicolumn{3}{c}{Memory Recall} & \multicolumn{3}{c}{Memory Reasoning} & 
\multicolumn{3}{c}{Memory Application} & \multirow{2}{*}{Overall} \\
\cmidrule(lr){4-6} \cmidrule(lr){7-9} \cmidrule(lr){10-12}
 & & & UPR & CRR & KR & MCR & RET & TR & TTL & CD & AR & \\
\midrule
\multirow{9}{*}{Text-based} & \multirow{3}{*}{Full (Text)} & D & 0.2747{\textcolor{blue}{\textbf{*}}} & 0.2378 & 0.4779{\textcolor{blue}{\textbf{*}}} & 0.2422{\textcolor{blue}{\textbf{*}}} & 0.2855{\textcolor{blue}{\textbf{*}}} & 0.3929{\textcolor{blue}{\textbf{*}}} & 0.3623 & 0.3009{\textcolor{blue}{\textbf{*}}} & 0.8456{\textcolor{blue}{\textbf{*}}} & 0.3496{\textcolor{blue}{\textbf{*}}} \\
                           & & M & 0.2155 & 0.3800{\textcolor{blue}{\textbf{*}}} & 0.2500 & 0.1413 & 0.2386 & 0.2000 & 0.3646{\textcolor{blue}{\textbf{*}}} & 0.1146 & 0.7188 & 0.3052 \\
                           & & D\&M & \dmcell 0.2694 & \dmcell 0.2471 & \dmcell 0.4520 & \dmcell 0.2351 & \dmcell 0.2821 & \dmcell 0.3715 & \dmcell 0.3626 & \dmcell 0.2820 & \dmcell 0.8339 & \dmcell 0.3464 \\
\cmidrule(lr){2-13}
                           & \multirow{3}{*}{NaiveRAG} & D & 0.5093 & 0.4445 & 0.4896{\textcolor{blue}{\textbf{*}}} & 0.3081 & 0.3269 & 0.5000 & 0.5428 & 0.3618{\textcolor{blue}{\textbf{*}}} & 0.8467{\textcolor{blue}{\textbf{*}}} & 0.4667 \\
                           & & M & 0.6048{\textcolor{blue}{\textbf{*}}} & 0.5104{\textcolor{blue}{\textbf{*}}} & 0.2500 & 0.3500{\textcolor{blue}{\textbf{*}}} & 0.4239{\textcolor{blue}{\textbf{*}}} & 0.5000 & 0.6400{\textcolor{blue}{\textbf{*}}} & 0.1250 & 0.8000 & 0.4933{\textcolor{blue}{\textbf{*}}} \\
                           & & D\&M & \dmcell 0.5181 & \dmcell 0.4489 & \dmcell 0.4563 & \dmcell 0.3111 & \dmcell 0.3340 & \dmcell \textbf{0.5000} & \dmcell 0.5542 & \dmcell 0.3377 & \dmcell 0.8424 & \dmcell 0.4569 \\
\cmidrule(lr){2-13}
                           & \multirow{3}{*}{A-Mem~\cite{xu2026amem}} & D & 0.6648 & 0.6070 & 0.5286{\textcolor{blue}{\textbf{*}}} & 0.4220 & 0.4515 & 0.4306{\textcolor{blue}{\textbf{*}}} & 0.5908 & 0.4014{\textcolor{blue}{\textbf{*}}} & 0.9356 & 0.5707 \\
                           & & M & 0.6694{\textcolor{blue}{\textbf{*}}} & 0.6700{\textcolor{blue}{\textbf{*}}} & 0.4000 & 0.4600{\textcolor{blue}{\textbf{*}}} & 0.5312{\textcolor{blue}{\textbf{*}}} & 0.3500 & 0.7083{\textcolor{blue}{\textbf{*}}} & 0.2400 & 1.0000{\textcolor{blue}{\textbf{*}}} & 0.5984{\textcolor{blue}{\textbf{*}}} \\
                           & & D\&M & \dmcell \textbf{0.6652} & \dmcell \textbf{0.6111} & \dmcell 0.5140 & \dmcell 0.4247 & \dmcell \textbf{0.4572} & \dmcell 0.4216 & \dmcell 0.6045 & \dmcell 0.3850 & \dmcell \textbf{0.9415} & \dmcell \textbf{0.5757} \\
\midrule
\multirow{9}{*}{Multi-modal} & \multirow{3}{*}{Full (MM)} & D & 0.3427{\textcolor{blue}{\textbf{*}}} & 0.3181{\textcolor{blue}{\textbf{*}}} & 0.5161{\textcolor{blue}{\textbf{*}}} & 0.3289{\textcolor{blue}{\textbf{*}}} & 0.3344{\textcolor{blue}{\textbf{*}}} & 0.3167 & 0.4681{\textcolor{blue}{\textbf{*}}} & 0.3073{\textcolor{blue}{\textbf{*}}} & 0.8107 & 0.4027{\textcolor{blue}{\textbf{*}}} \\
                           & & M & 0.2903 & 0.2800 & 0.4500 & 0.2708 & 0.2826 & 0.3500{\textcolor{blue}{\textbf{*}}} & 0.4583 & 0.1739 & 0.8152{\textcolor{blue}{\textbf{*}}} & 0.3648 \\
                           & & D\&M & \dmcell 0.3380 & \dmcell 0.3156 & \dmcell 0.5086 & \dmcell 0.3248 & \dmcell 0.3307 & \dmcell 0.3204 & \dmcell 0.4670 & \dmcell 0.2938 & \dmcell 0.8111 & \dmcell 0.3988 \\
\cmidrule(lr){2-13}
                           & \multirow{3}{*}{MuRAG~\cite{chen-etal-2022-murag}} & D & 0.6312 & 0.5216 & 0.5096 & 0.4407{\textcolor{blue}{\textbf{*}}} & 0.4442 & 0.3833{\textcolor{blue}{\textbf{*}}} & 0.6052 & 0.3939{\textcolor{blue}{\textbf{*}}} & 0.9002 & 0.5496 \\
                           & & M & 0.6694{\textcolor{blue}{\textbf{*}}} & 0.6900{\textcolor{blue}{\textbf{*}}} & 0.6500{\textcolor{blue}{\textbf{*}}} & 0.3100 & 0.4762{\textcolor{blue}{\textbf{*}}} & 0.2000 & 0.7000{\textcolor{blue}{\textbf{*}}} & 0.2400 & 1.0000{\textcolor{blue}{\textbf{*}}} & 0.5757{\textcolor{blue}{\textbf{*}}} \\
                           & & D\&M & \dmcell 0.6346 & \dmcell 0.5326 & \dmcell \textbf{0.5255} & \dmcell \textbf{0.4315} & \dmcell 0.4465 & \dmcell 0.3629 & \dmcell \textbf{0.6162} & \dmcell 0.3782 & \dmcell 0.9094 & \dmcell 0.5527 \\
\cmidrule(lr){2-13}
                           & \multirow{3}{*}{NGM~\cite{fisher2025neural}} & D & 0.5119{\textcolor{blue}{\textbf{*}}} & 0.4506{\textcolor{blue}{\textbf{*}}} & 0.4872{\textcolor{blue}{\textbf{*}}} & 0.3576 & 0.3998 & 0.4562{\textcolor{blue}{\textbf{*}}} & 0.5635 & 0.4509{\textcolor{blue}{\textbf{*}}} & 0.9072 & 0.4946 \\
                           & & M & 0.5081 & 0.3700 & 0.4500 & 0.4300{\textcolor{blue}{\textbf{*}}} & 0.4271{\textcolor{blue}{\textbf{*}}} & 0.4000 & 0.7000{\textcolor{blue}{\textbf{*}}} & 0.2400 & 1.0000{\textcolor{blue}{\textbf{*}}} & 0.5172{\textcolor{blue}{\textbf{*}}} \\
                           & & D\&M & \dmcell 0.5116 & \dmcell 0.4454 & \dmcell 0.4830 & \dmcell 0.3627 & \dmcell 0.4017 & \dmcell 0.4500 & \dmcell 0.5794 & \dmcell \textbf{0.4295} & \dmcell 0.9157 & \dmcell 0.5049 \\
\bottomrule
\end{tabular}
\end{table}

\subsection{Task Design}

To systematically evaluate these capabilities, we design a hierarchical taxonomy of nine task types, organized into three categories. Figure~\ref{fig:problem} shows example questions for all task types.

\noindent \textbf{Memory Recall.}
This category evaluates whether models can retrieve explicitly presented multimodal information. (1) \textbf{Unimodal Precise Recall (UPR)}: Given a query $q$, the model retrieves information from a single modality $x_{t,i}$ or $v_{t,i}$. (2) \textbf{Cross-modal Related Retrieval (CRR)}: The model retrieves aligned content across modalities, i.e., mapping text $x_{t,i}$ to image $v_{t,i}$ or vice versa. (3) \textbf{Knowledge Resolution (KR)}: Given multi-session dialogues $S = (s_1, \dots, s_T)$ with updated information across sessions, the model retrieves the currently correct information from memory $\mathcal{M}_T$.

\noindent \textbf{Memory Reasoning.}
This category evaluates reasoning over multimodal information across time and participants. (1) \textbf{Temporal Reasoning (TR)}: The model orders events across sessions using timestamps $\tau_t$ and utterance positions. (2) \textbf{Multimodal Causal Reasoning (MCR)}: The model infers causal relations between textual content $x_{t,i}$ and visual content $v_{t',j}$ across sessions and speakers. (3) \textbf{Reference \& Evolution Tracking (RET)}: The model resolves references and tracks entity evolution across sessions $s_t$ and speakers $p_{t,i}$.

\noindent \textbf{Memory Application.}
This category evaluates how models apply and update memory during inference. (1) \textbf{Test-Time Learning (TTL)}: The model adapts to new scenarios at inference time by using memory $\mathcal{M}_T$. (2) \textbf{Conflict Detection (CD)}: The model detects whether a new statement contradicts $\mathcal{M}_T$. (3) \textbf{Answer Refusal (AR)}: The model refuses to answer when information is absent from $\mathcal{M}_T$ or cannot be inferred.

\section{Experiment}
\begin{table}[htbp]
\centering
\caption{Weighted average (D\&M) performance of different methods across all categories. Metrics: P=Precision, R=Recall, F1=F1-score, B=BLEU-1. Results are from GPT-4.1-nano~\cite{openai2024gpt4technicalreport} with top-5 retrieval. Bold values indicate the best performance among the six methods within each metric column for the given category. Additional results on other backbone models are reported in Appendix~\ref{app:sec4.2}.}
\label{tab:lexical_metrics_Gpt-4.1-Nano}
\fontsize{7pt}{8.5pt}\selectfont
\setlength{\tabcolsep}{6pt}
\renewcommand{\arraystretch}{1.0}
\begin{tabular}{@{} l l l *{10}{c} @{}}
\toprule
\multirow{2}{*}{Category} & \multirow{2}{*}{Method} & \multirow{2}{*}{Metrics} & 
\multicolumn{3}{c}{Memory Recall} & \multicolumn{3}{c}{Memory Reasoning} & 
\multicolumn{3}{c}{Memory Application} & \multirow{2}{*}{Overall} \\
\cmidrule(lr){4-6} \cmidrule(lr){7-9} \cmidrule(lr){10-12}
 & & & UPR & CRR & KR & MCR & RET & TR & TTL & CD & AR & \\
\midrule
\multirow{12}{*}{Text-based} & \multirow{4}{*}{Full (Text)} & P & 0.1412 & 0.1074 & 0.3290 & 0.1197 & 0.1317 & 0.4469 & 0.1184 & 0.0650 & 0.8279 & 0.2394 \\
 & & R & 0.2111 & 0.2212 & 0.3120 & 0.1950 & 0.2239 & 0.4802 & 0.2260 & 0.0521 & 0.8255 & 0.2637 \\
 & & F1 & 0.1479 & 0.1277 & 0.3071 & 0.1346 & 0.1470 & 0.3997 & 0.1381 & 0.0550 & 0.8215 & 0.2391 \\
 & & B & 0.1153 & 0.0965 & 0.2461 & 0.1069 & 0.1083 & 0.2830 & 0.1086 & 0.0489 & 0.8172 & 0.2299 \\
\cmidrule(lr){2-13}
 & \multirow{4}{*}{NaiveRAG} & P & \textbf{0.3136} & \textbf{0.2264} & 0.3249 & 0.1330 & 0.1364 & 0.6112 & \textbf{0.1917} & \textbf{0.2378} & 0.8412 & \textbf{0.3082} \\
 & & R & 0.3605 & 0.2967 & 0.2429 & 0.1843 & 0.1682 & 0.3773 & 0.3138 & \textbf{0.2158} & 0.8353 & 0.3042 \\
 & & F1 & \textbf{0.3041} & \textbf{0.2383} & 0.2601 & 0.1420 & 0.1299 & 0.4386 & 0.2119 & \textbf{0.2194} & 0.8330 & \textbf{0.2999} \\
 & & B & \textbf{0.2575} & \textbf{0.2045} & 0.1656 & 0.1145 & 0.0954 & 0.2577 & 0.1754 & \textbf{0.2112} & 0.8309 & \textbf{0.2841} \\
\cmidrule(lr){2-13}
 & \multirow{4}{*}{A-Mem~\cite{xu2026amem}} & P & 0.1384 & 0.0942 & 0.3296 & 0.0958 & 0.1103 & 0.2258 & 0.0767 & 0.1036 & 0.8834 & 0.2206 \\
 & & R & \textbf{0.4544} & \textbf{0.4390} & \textbf{0.3988} & \textbf{0.3712} & \textbf{0.3657} & \textbf{0.6550} & \textbf{0.4325} & 0.0869 & \textbf{0.8979} & \textbf{0.4215} \\
 & & F1 & 0.1887 & 0.1410 & \textbf{0.3483} & 0.1380 & 0.1549 & 0.2895 & 0.1251 & 0.0887 & 0.8748 & 0.2364 \\
 & & B & 0.1257 & 0.0828 & 0.2903 & 0.0847 & 0.1020 & 0.2006 & 0.0795 & 0.0027 & 0.8690 & 0.2120 \\
\midrule
\multirow{12}{*}{Multi-modal} & \multirow{4}{*}{Full (MM)} & P & 0.1244 & 0.0865 & 0.3053 & 0.1045 & 0.1071 & 0.3612 & 0.0968 & 0.1298 & 0.7850 & 0.2225 \\
 & & R & 0.2787 & 0.2558 & 0.3557 & 0.2498 & 0.2671 & 0.4891 & 0.2991 & 0.1180 & 0.7906 & 0.3034 \\
 & & F1 & 0.1458 & 0.1154 & 0.3148 & 0.1343 & 0.1372 & 0.3395 & 0.1275 & 0.1175 & 0.7854 & 0.2259 \\
 & & B & 0.1069 & 0.0793 & 0.2575 & 0.0953 & 0.0955 & 0.2408 & 0.0843 & 0.1114 & 0.7795 & 0.2137 \\
\cmidrule(lr){2-13}
 & \multirow{4}{*}{MuRAG~\cite{chen-etal-2022-murag}} & P & 0.1747 & 0.0994 & 0.3513 & 0.1152 & 0.1299 & 0.3984 & 0.1142 & 0.1101 & 0.8856 & 0.2601 \\
 & & R & 0.4063 & 0.3120 & 0.3581 & 0.2923 & 0.3194 & 0.5559 & 0.3672 & 0.1067 & 0.8898 & 0.3443 \\
 & & F1 & 0.2179 & 0.1349 & 0.3451 & 0.1472 & 0.1661 & 0.3657 & 0.1541 & 0.0999 & 0.8749 & 0.2738 \\
 & & B & 0.1529 & 0.0861 & \textbf{0.2928} & 0.1049 & 0.1150 & 0.2280 & 0.0951 & 0.0118 & 0.8702 & 0.2453 \\
\cmidrule(lr){2-13}
 & \multirow{4}{*}{NGM~\cite{fisher2025neural}} & P & 0.2416 & 0.1412 & \textbf{0.4053} & \textbf{0.1498} & \textbf{0.1662} & \textbf{0.6711} & 0.1844 & 0.0780 & \textbf{0.8933} & 0.2858 \\
 & & R & 0.3471 & 0.2964 & 0.3110 & 0.2500 & 0.2692 & 0.5189 & 0.3552 & 0.0681 & 0.8862 & 0.3243 \\
 & & F1 & 0.2537 & 0.1710 & 0.3438 & \textbf{0.1709} & \textbf{0.1853} & \textbf{0.5255} & \textbf{0.2157} & 0.0712 & \textbf{0.8804} & 0.2804 \\
 & & B & 0.1737 & 0.1196 & 0.2660 & \textbf{0.1292} & \textbf{0.1393} & \textbf{0.3287} & 0.1457 & 0.0000 & \textbf{0.8747} & 0.2629 \\
\bottomrule
\end{tabular}
\end{table}
\subsection{Experimental Setup}

\noindent \textbf{Backbone and Memory Method.} We conduct a comprehensive evaluation of both text-based and multimodal memory methods. Specifically, text-based methods include Full Memory (Text), NaiveRAG, and A-Mem~\cite{xu2026amem}, while multimodal memory methods include Full Memory (Multimodal), MuRAG~\cite{chen-etal-2022-murag}, and NGM~\cite{fisher2025neural}. More detailed explanations of memory methods is shown in Appendix~\ref{app:sec2}. All methods are evaluated using multimodal large language models (MLLMs) as the Backbone, including the Qwen2.5-VL family (3B and 7B instruct variants)~\cite{bai2025qwen25vltechnicalreport} and GPT-4.1-Nano~\cite{openai2024gpt4technicalreport}. For methods that require retrieval, we adopt a dense retriever with a default top-$K=5$. To enable a fair comparison between text-based and multimodal memory systems, we augment textual memory methods with high-quality image captions generated by GPT-4o~\cite{openai2024gpt4ocard}. More implement details can be found in Appendix~\ref{app:sec3.1}.

\noindent \textbf{Evaluation Metrics.} To systematically assess agent performance, we adopt an LLM-as-Judge approach as our primary evaluation metric. Specifically, we employ GPT-4o-mini as a zero-shot evaluator to score each model response against the ground truth. We validated this approach by measuring agreement with human judgments on a 200-sample subset, achieving Cohen's $\kappa = 0.84$ which indicates near-perfect agreement~\cite{landis1977measurement}. As a complement, we also report traditional lexical metrics, including precision, recall, F1 score, and BLEU-1. The details can be found in Appendix~\ref{app:sec3.3}.

\subsection{Experimental Results}
Table~\ref{tab:dyadic_multiparty_combined} and Table~\ref{tab:lexical_metrics_Gpt-4.1-Nano} report the overall performance evaluated by LLM-as-Judge and lexical metrics, respectively. Overall performance remains low, with the best weighted average LLM-as-Judge score reaching only 0.5757 (A-Mem~\cite{xu2026amem}). Combining semantic correctness and lexical fidelity, we identify four major bottlenecks in current memory systems. \textbf{(1) Cross-modal alignment remains challenging.} A consistent gap exists between Unimodal Precise Recall (UPR) and Cross-modal Related Retrieval (CRR). For example, MuRAG~\cite{chen-etal-2022-murag} drops from 0.6346 to 0.5326 in LLM-as-Judge scores, with a similar lexical gap (recall: 0.4063 vs. 0.3120). \textbf{(2) Weak distractor filtering despite successful retrieval.} A large recall--precision gap is observed across methods; for instance, A-Mem~\cite{xu2026amem} achieves 0.4215 recall but only 0.2206 precision, indicating difficulty filtering noisy multiple participants' information while agents can successfully retrieve relevant history. \textbf{(3) Limited causal reasoning and adaptation to human referential conventions.} Reasoning tasks, especially Multimodal Causal Reasoning (MCR) and Reference \& Evolution Tracking (RET), consistently show the lowest scores. Moreover, the near-zero BLEU-1 scores in these tasks (Table~\ref{tab:lexical_metrics_Gpt-4.1-Nano}) indicate that models rarely reproduce the precise factual phrasing needed to connect distributed evidence, particularly under human preferences for implicit reference. \textbf{(4) Poor robustness to conflicting information.} Conflict Detection (CD) remains particularly difficult, with near-zero lexical precision and recall (e.g., A-Mem CD recall: 0.0869), highlighting the inability to resolve contradictions in human--human interactions.

\begin{figure}[htbp]
\centering
\includegraphics[width=\textwidth]{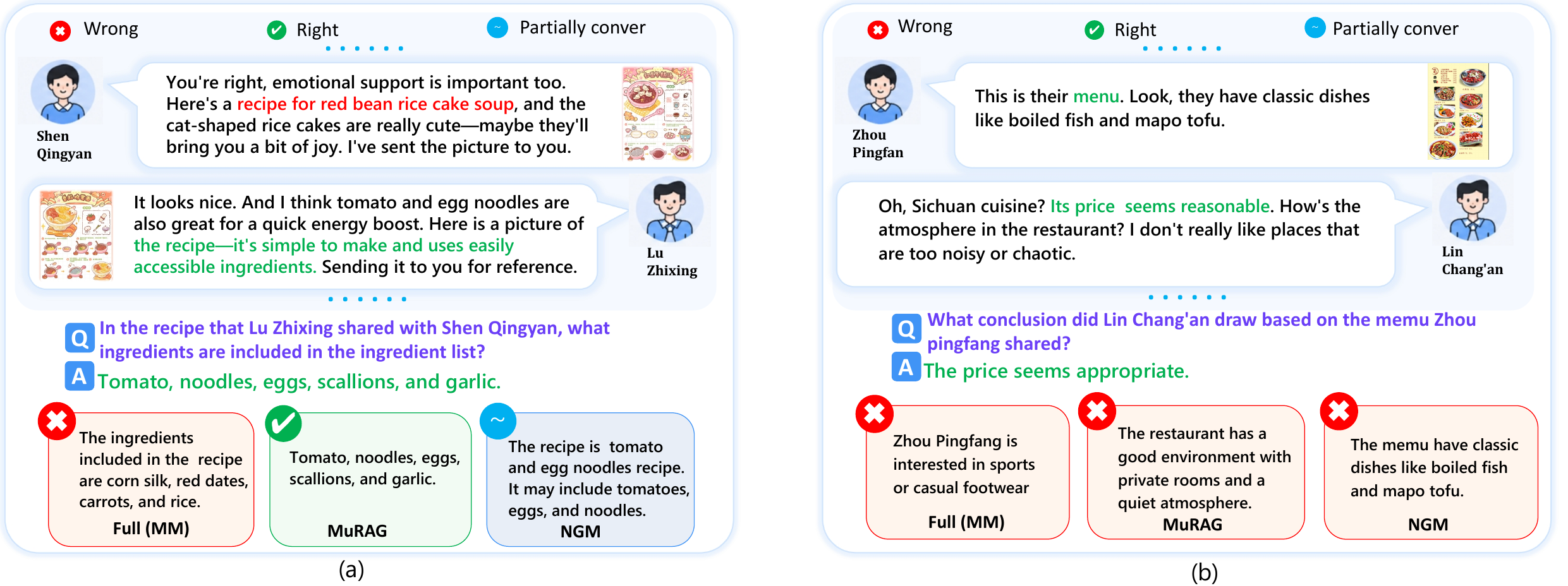}
\caption{Case studies of multimodal conversational reasoning. (a) Identifying ingredients in Lu Zhixing’s recipe. (b) Inferring Lin Chang'an’s conclusion based on a shared menu.}
\label{fig:case_study}
\end{figure}

\noindent \textbf{Impact of Interaction Structure: Dyadic vs. Multi-party.} To understand how interaction structure affects agent memory, we compare performance across dyadic and multi-party settings (Table~\ref{tab:dyadic_multiparty_combined}). Dyadic dialogues span longer time horizons with more sessions (avg. 14.2 sessions), whereas multi-party dialogues contain denser interactions within fewer sessions (avg. 70.5 rounds/session and 5.0 sessions). This difference leads to complementary performance patterns. Consistency-oriented tasks such as Knowledge Resolution (KR) and Conflict Detection (CD) are substantially harder in multi-party settings due to contradictory signals from multiple speakers. For example, NaiveRAG's KR score drops from 0.4896 in dyadic to 0.2500 in multi-party dialogues. In contrast, tasks benefiting from concentrated contextual evidence, such as Cross-modal Related Retrieval (CRR) and Test-Time Learning (TTL), achieve comparable or higher performance in multi-party settings. Moreover, experiments with larger backbones (Qwen2.5-VL-7B-Instruct~\cite{bai2025qwen25vltechnicalreport}, detailed in Appendix~\ref{app:sec4.1}) show that parameter scaling alone does not eliminate this gap, indicating that current memory mechanisms remain insufficiently robust to diverse interaction structures.

\begin{table}[htbp]
\centering
\fontsize{7pt}{9pt}\selectfont
\setlength{\tabcolsep}{4pt}
\renewcommand{\arraystretch}{1.0}
\caption{
Efficiency comparison across methods. Storage is measured per session (s/sess), while retrieval and generation are measured per query (s/q).
}
\label{tab:efficiency_analysis}

\begin{tabular}{l c c c}
\toprule
\textbf{Method} & \textbf{Storage} (s/sess) $\downarrow$ & \textbf{Retrieval} (s/q) $\downarrow$ & \textbf{Answer} (s/q) $\downarrow$ \\
\midrule
Full (Text)   & 0.0015 & 0.1566 & 17.99 \\
NaiveRAG      & 0.6946 & 1.3710 & 10.06 \\
A-Mem~\cite{xu2026amem}         & 351.08 & 0.0248 & 4.57 \\
\midrule
Full (MM)     & 0.0009 & 0.3597 & 26.09 \\
MuRAG~\cite{chen-etal-2022-murag}         & 9.861  & 1.4674 & 12.64 \\
NGM~\cite{fisher2025neural}           & 6.529  & 0.7734 & 4.33 \\
\bottomrule
\end{tabular}

\end{table}

\noindent \textbf{Efficiency Trade-offs.} Beyond accuracy, tracking multimodal human--human interactions imposes substantial computational burdens. Table~\ref{tab:efficiency_analysis} reveals a clear trade-off between storage and inference latency. Full-memory methods introduce minimal storage overhead but suffer from severe inference latency, especially with multimodal inputs (17.99 s/q for Full (Text) vs. 26.09 s/q for Full (MM)). In contrast, agentic memory systems such as A-Mem~\cite{xu2026amem} reduce inference latency but incur high memory construction costs (351.08 s/session). These results highlight the need for lightweight memory compression paradigms for multimodal observer agents. Additional retriever analysis is provided in Appendix~\ref{app:sec4.3}.

\subsection{In-depth Analysis and Case Study}
\noindent \textbf{In-depth Analysis.} To move beyond aggregated metrics, we manually analyze 100 failed cross-modal and reasoning instances from three multimodal memory methods, categorizing them into four archetypes (Table~\ref{tab:error_distribution}). The errors are highly concentrated in two major failure modes. Modal misalignment accounts for 44\%--46\% of cases, showing that current systems struggle to ground textual content in visual evidence. Speaker-related errors account for 32\%--35\%, highlighting difficulties in maintaining correct participant attribution and resolving human referential expressions in human--human interactions.

\begin{table}[htbp]
\centering
\fontsize{7pt}{9pt}\selectfont
\setlength{\tabcolsep}{6pt}      
\renewcommand{\arraystretch}{1.2} 
\caption{
Distribution of error types for representative multimodal methods. 
Error archetypes are defined as follows: 
\textit{Modal Misalignment}: failing to align text with image; 
\textit{Speaker-related Errors}: wrong person attribution or failing to follow human referential preference; 
\textit{Temporal Confusion}: using outdated information or reversing event order; 
\textit{Other / Hallucination}: remaining failure cases not covered by the above categories.
}
\label{tab:error_distribution}
\begin{tabular}{lccc}
\toprule
\textbf{Error Archetype} & 
\makecell[c]{\textbf{Full (MM)}} & 
\makecell[c]{\textbf{MuRAG} \\ \cite{chen-etal-2022-murag}} & 
\makecell[c]{\textbf{NGM} \\ \cite{fisher2025neural}} \\
\midrule
Modal Misalignment  & 48\% & 44\% & 46\% \\
Speaker-related Errors & 37\% & 35\% & 32\% \\
Temporal Confusion   & 15\% & 16\% & 9\% \\
Other / Hallucination & 5\% & 5\% & 6\% \\
\bottomrule
\end{tabular}
\end{table}

\noindent \textbf{Case study.} Two representative cases (Figure~\ref{fig:case_study}) illustrate how dominant failure modes manifest in practice. Case (a) focuses on ingredient identification from Lu Zhixing’s recipe, requiring fine-grained visual grounding and correct speaker–image alignment; NGM~\cite{fisher2025neural} exhibits modal misalignment by ignoring the image, while Full (MM) fails by misattributing the recipe image to the wrong speaker. Case (b) infers Lin Chang'an’s conclusion drawn from a shared menu, which hinges on causal reasoning; NGM~\cite{fisher2025neural} misattributing the conclusion to the wrong speaker, MuRAG~\cite{chen-etal-2022-murag} fails to align visual evidence with the textual conclusion, mistakenly presenting follow-up results instead.. Full (MM) is distracted by other visual information, causing it to provide a completely irrelevant conclusion.

\section{Conclusion}

We introduce H2HMem, a benchmark for evaluating multimodal memory in LLM agents within human--human interactions, providing a unified framework for assessing memory recall, reasoning, and application. Experiments show that current methods can retrieve relevant information but remain weak at integrating it. They can recall fragments — images, facts, statements — but fail to align visual evidence with text, attribute information to the correct speaker across sessions, or resolve contradictions from multiple sources. These failures persist across dyadic and multi-party settings, revealing that in multimodal human--human interactions, remembering fragments is not enough; agents must reconstruct multimodal coherent memory from distributed human communications.

\bibliographystyle{unsrt}

\appendix

\section{Dataset Details}
\label{app:sec1}

\begin{table}[bp]
\centering
\small
\setlength{\tabcolsep}{3.5pt}
\renewcommand{\arraystretch}{1.0}
\caption{Statistics of dyadic and multi-party conversations in the H2HMem dataset.}
\label{tab:all_stats}
\begin{tabular}{l c c c c l}
\toprule
\multicolumn{6}{c}{\textbf{Dyadic Conversations}} \\
\midrule
\textbf{Dialogue} & \textbf{Sessions} & \textbf{Rounds} & \textbf{Images} & \textbf{\#Topics} & \textbf{Topics} \\
\midrule
1  & 11 & 215 & 35 & 5 & food, health, news, pet, shopping, sport \\
2  & 11 & 215 & 36 & 4 & food, health, movie, pet, shopping \\
3  & 15 & 253 & 54 & 5 & annual\_summary, entertain, food, news, pet, shopping \\
4  & 15 & 280 & 61 & 6 & entertain, food, health, movie, pet, shopping, travel \\
5  & 14 & 263 & 49 & 7 & annual\_summary, entertain, food, health, movie, news, pet, shopping \\
6  & 11 & 214 & 37 & 4 & food, health, pet, sport, travel \\
7  & 12 & 221 & 41 & 4 & food, health, pet, shopping, travel \\
8  & 14 & 254 & 51 & 6 & annual\_summary, food, health, movie, pet, shopping, travel \\
9  & 16 & 314 & 58 & 6 & annual\_summary, entertain, food, health, news, pet, shopping \\
10 & 16 & 293 & 44 & 7 & annual\_summary, entertain, food, health, movie, news, pet, shopping \\
11 & 13 & 245 & 36 & 6 & annual\_summary, entertain, food, health, news, pet, shopping \\
12 & 16 & 278 & 46 & 6 & annual\_summary, entertain, health, movie, news, pet, shopping \\
13 & 15 & 290 & 53 & 5 & annual\_summary, entertain, food, health, pet, shopping \\
14 & 16 & 297 & 55 & 6 & annual\_summary, entertain, health, news, pet, shopping \\
15 & 15 & 275 & 54 & 4 & food, health, pet, shopping, sport \\
16 & 15 & 286 & 49 & 4 & food, health, movie, pet, shopping \\
17 & 15 & 273 & 44 & 4 & entertain, health, news, pet \\
18 & 16 & 310 & 48 & 4 & food, health, pet, shopping, sport \\
19 & 16 & 308 & 53 & 4 & food, health, movie, pet, shopping \\
20 & 12 & 232 & 47 & 5 & annual\_summary, food, entertain, health, pet, sport \\
\midrule
\multicolumn{6}{c}{\textbf{Multi-party Conversations}} \\
\midrule
\textbf{Dialogue} & \textbf{Sessions} & \textbf{Rounds} & \textbf{Images} & \textbf{\#Topics} & \textbf{Topics} \\
\midrule
1 & 5 & 358 & 70 & 3 & food, pet, shopping \\
2 & 5 & 352 & 70 & 3 & sport, shopping, health \\
3 & 5 & 358 & 69 & 4 & movie, entertain, shopping, travel \\
4 & 5 & 331 & 70 & 3 & shopping, food, entertain \\
5 & 5 & 363 & 70 & 3 & travel, shopping, work \\
\midrule
\end{tabular}
\end{table}
We provide more detailed dialogue statistics and additional details on key components of our data construction pipeline that are not fully elaborated in the main paper.

\subsection{Conversation Data Statistics}
\label{app:sec1.1}
Table~\ref{tab:all_stats} provides per-dialogue statistics for the 20 dyadic and 5 multi-party conversations in H2HMem. Dyadic dialogues contain 11–16 sessions per conversation (median 15) and 214–314 rounds, accompanied by 35–61 images. Multi-party dialogues are designed with fewer sessions (5 each) but much denser rounds (331–363 per dialogue) and a higher image count (69–70 per dialogue), reflecting the “intensive single-session” characteristic discussed in the main paper. The topic coverage is also broader in dyadic settings (4–7 topics per dialogue, covering food, health, pet, shopping, travel, news, movie, entertainment, sport, and annual\_summary) compared to multi-party dialogues (3–4 topics, with shopping present in all five). The time span for a single conversation is limited to one year. This detailed breakdown confirms the structural and topical diversity of our dataset, which is essential for evaluating memory under realistic human–human interaction conditions. All dialogues and question-answer pairs are in English.

\subsection{Persona Schema}
\label{app:sec1.2}
In the main paper, we mention that participant profiles follow a structured schema. Table~\ref{tab:persona_schema} presents the complete schema and a concrete example from our dataset. For dyadic dialogues, we generate two participant profiles per dialogue. For multi‑party dialogues, we generate four to six participant profiles per dialogue.

\begin{table}[htbp]
\centering
\caption{Complete participant profile schema and example.}
\label{tab:persona_schema}
\small
\setlength{\tabcolsep}{3pt}
\renewcommand{\arraystretch}{1.2}
\begin{tabular}{p{2.8cm}|p{3cm}|p{8cm}}
\toprule
\textbf{Field} & \textbf{Description} & \textbf{Example} \\
\midrule
name & Full name & Zhao Xiaotang \\
age & Age & 26 \\
gender & Gender & Female \\
profession & Job role & Social work organization project specialist / community volunteer \\
title & Position & Project Supervisor \\
specialty & Areas of expertise & Community building, vulnerable group assistance, resource networking, public welfare project management \\
personality\_traits & Character traits & Warm-hearted, practical and down-to-earth, strong organizational skills, slightly prone to over-worrying \\
core\_values & Core values & Mutual assistance and sharing, community belonging, pragmatism, human warmth \\
fears & Anxieties or concerns & Fear of community indifference, worry about resource waste, fatigue from excessive sense of responsibility \\
motivations & Intrinsic drivers & Build mutual aid networks, improve community environment, help those in need, facilitate resource flow \\
background & Life background & From an ordinary working-class family, grew up in a company dormitory community with a strong sense of collective belonging \\
education & Educational background & B.A. in Social Work, East China Normal University \\
relationships & Social connections & Single, lives with parents, a "well-known figure" in the community, feeds three stray cats in the neighborhood \\
\bottomrule
\end{tabular}
\end{table}

\subsection{Outline Generation}
\label{app:sec1.3}
The main paper describes that we generate session-level outlines to guide dialogue generation. Table~\ref{tab:outline_example} shows a complete outline example. For dyadic dialogues, we prompt the language model to sample 4–6 topics, and for each topic, generate 3–4 session-level outlines, ensuring a total of no fewer than 10 outlines per dialogue. For multi-party dialogues, we prompt the model to sample 3–4 topics, with 1–2 session-level outlines per topic, keeping the total number of outlines no more than 5 per dialogue.

\begin{table}[htbp]
\centering
\caption{Example session outline.}
\label{tab:outline_example}
\small
\setlength{\tabcolsep}{2.5pt}
\renewcommand{\arraystretch}{1.2}
\begin{tabular}{p{2.6cm}|p{2.6cm}|p{8.5cm}}
\toprule
\textbf{Field} & \textbf{Description} & \textbf{Content} \\
\midrule
session\_title & Brief title of the session & DIY Implementation and Neutering Plan Emergence \\
theme & High-level topic category & Pets - Stray Cat TNR Project \\
sequence\_number & Order of this session in the scenario & 2 \\
timeline\_date & In-story date & 2024-03-15 \\
timeline\_remark & Temporal relationship to previous session & One week later, cat shelter completed \\
core\_anchor & Key event or image trigger that drives the session & Zhao Xiaotang sends a photo of the finished foam box cat shelter (already placed, with mother cat and kittens inside) and thanks Li Yifan for his "ingenious design." She also expresses concern: "This solves the immediate crisis, but the mother cat will go into heat again soon. This can't go on indefinitely." \\
scenario\_flow & Step-by-step description of how the dialogue should unfold & Li Yifan agrees and explains that TNR (Trap-Neuter-Return) is the most humane long-term solution internationally. He begins researching whether local animal protection organizations or pet hospitals offer discounted spay/neuter programs for strays, and creates a comparison table of costs, appointment procedures, and post-operative care. Zhao Xiaotang starts posting initiatives in the community group to see if neighbors are willing to share neutering costs or provide temporary post-surgery housing. \\
end\_state & How the session concludes & The immediate crisis resolved, shifting focus toward a fundamental solution, beginning community mobilization and resource research. \\
character\_states & How each participant's role or mindset evolves during the session & Zhao Xiaotang: Transitioning from hands-on rescue to systematic solution thinking, activating community organizing capabilities. \\
 & & Li Yifan: Shifting from technical support to solution research and cost analysis, providing decision-making basis. \\
key\_constraints & Specific guidelines for dialogue generation (e.g., what to avoid or emphasize) & The cat shelter photo should not be praised for "looking good," but rather describe practical effects such as "the inside of the foam box is lined with aluminum insulation, with ventilation holes on top" and "the mother cat has been willing to take the kittens inside." \\
\bottomrule
\end{tabular}
\end{table}

\subsection{Human Annotation Protocol}
\label{app:sec1.4}
The main paper mentions two rounds of human verification. Below we detail the specific criteria, annotator training, and quality control procedures. Six undergraduate annotators participated, all native speakers of the dialogue language and familiar with multimodal data annotation.

\subsubsection{Image Refinement}
\label{app:sec1.4.1}
For each dialogue, the pipeline automatically retrieved or generated candidate images based on textual triggers. Six undergraduate annotators then reviewed every image together with its surrounding conversation. They focused on three aspects. First, the visual content had to match the triggering utterance exactly — for example, if a participant said “Look at this X-ray”, the image had to actually show an X‑ray, not a generic medical illustration. Second, the image quality needed to be sufficient for captioning and human interpretation: a resolution of at least 224×224 pixels, and no heavy blur, compression artifacts, or abstract drawings. Third, the image had to be topically appropriate and free of offensive or misleading content; in dyadic or multi‑party settings, it also had to be plausible given the conversation topic (e.g., a vacation photo for travel discussions, a budget chart for a meeting). When an image failed any of these checks, annotators could replace it with a manually retrieved alternative from a stock photo library, edit the image (e.g., cropping or annotating), or — as a last resort — request a complete re‑generation of the corresponding dialogue segment. The entire process of image refinement took approximately 80 person‑hours.

\subsubsection{QA Validation}
\label{app:sec1.4.2}
The same annotators then examined each automatically generated question–answer pair. They checked whether the answer could be uniquely derived from the dialogue history and its images — if two different dates were mentioned without resolution, the pair was discarded or rewritten. They also ensured that the question itself was unambiguous, containing no vague references like “What about that thing?” that the context could not resolve; special attention was paid to anaphora and deixis. Finally, each QA pair was assigned to one of the nine task types (UPR, CRR, KR, MCR, RET, TR, TTL, CD, AR), and the annotators judged whether the required reasoning indeed matched the intended difficulty level (recall, reasoning, or application). Mismatched pairs were sent back for re‑assignment or re‑generation. The validation process took approximately 40 person‑hours. Each pair was refined independently by two annotators, with cross‑checking to catch individual errors.

\subsubsection{Inter-Annotator Agreement}
To ensure consistency, we measured Fleiss’ $\kappa$ for each round on a subset of 10\% of the data. For image refinement, the agreement was $\kappa = 0.83$ (substantial); for QA validation, $\kappa = 0.79$ (substantial). Disagreements were resolved through discussion led by a senior researcher, who made the final decision when consensus could not be reached.

\subsubsection{Time and Effort}
The entire annotation process took approximately 120 person‑hours: 80 hours for image refinement (including manual replacements) and 40 hours for QA validation. Each round was performed independently by two annotators, with cross‑checking to catch individual errors.

\subsubsection{Recruitment and Payment}
Six undergraduate annotators were recruited from the authors' institution. They were compensated with a small stipend of \$1 per hour, which is commensurate with local student wages.

\subsubsection{Consent}
All annotators provided informed consent prior to participation. They were informed that the data would be used solely for academic research purposes.

\section{Baseline Models}
\begin{figure}[htbp]
\centering
\begingroup
\linespread{1.0}\selectfont
\begin{tcolorbox}[
    colback=gray!5,
    colframe=black,
    title=Memory Evaluating Prompt Template,
    width=1.0\textwidth,
    arc=3mm,
    boxrule=0.5pt,
    left=3mm,
    right=3mm,
    top=2.5mm,
    bottom=2.5mm,
    fonttitle=\bfseries
]

You are a memory evaluating system. \{instruction\}

IMPORTANT:
1. Provide only the answer without any reasoning process. Give the answer directly in English.
2. Keep your answer within 100 words. Short and concise answers are acceptable.
3. Answer in English. This is a strict requirement. Do not answer in any other language.

\{context\_section\}

\{context\_note\}

Question: \{question\}

\{format\_requirement\}

Examples:
Question: What is the cat's name?
Correct answer: Almond

Incorrect answer example (DO NOT answer like this):
We need answer: cat name is Almond because...

\end{tcolorbox}
\endgroup
\caption{Memory evaluating prompt template.}
\label{fig:eval_prompt}
\end{figure}
\label{app:sec2}
We evaluate six baseline methods, categorized as text-based or multimodal. 

\subsection{Text-based Methods (with Image Captions)}
For text-based memory methods, raw images are converted into image captions using GPT-4o~\cite{openai2024gpt4ocard} before being stored in memory.

\noindent \textbf{Full Memory (Text).} It includes all session transcripts and image captions in textual form as part of the context, and truncates the input according to the context token limit. No retrieval or compression is applied.

\noindent \textbf{NaiveRAG.} It splits the conversation history into chunks. Each chunk is encoded into a semantic vector. At query time, it retrieves the top-\(K=5\) most relevant chunks based on vector similarity and concatenates them with the query.

\noindent \textbf{A-Mem~\cite{xu2026amem}.} It constructs structured memory episodes during preprocessing, where each episode summarizes a coherent discourse unit (e.g., 5--10 dialogue turns). New memories autonomously establish links to related past memories. At inference time, it retrieves relevant episodes and performs memory consolidation to support long-term reasoning.

\subsection{Multimodal Methods}
For multimodal memory methods, raw images are stored and retrieved directly without conversion to text.

\noindent \textbf{Full Memory (Multimodal).} It includes all multimodal memory information (interleaved text and raw images) as context, estimates the token consumption of images using predefined token costs, and truncates the input according to the context token limit.

\noindent \textbf{MuRAG~\cite{chen-etal-2022-murag}.} It uses a dense multimodal retriever that encodes both queries and memory entries into a shared embedding space using a joint vision-language encoder. At inference time, it performs maximum inner product search over an external memory to retrieve the top-\(K=5\) most relevant multimodal passages, which are then used to augment generation.

\noindent \textbf{NGM~\cite{fisher2025neural}.} It proposes Neural Generative Memory, which maintains a compressed latent representation of the conversation history. Memory is updated incrementally as new dialogue turns arrive, enabling efficient long-term memorization without explicit retrieval at each step.

For methods without publicly available implementation code, we re-implement them based on the methodological descriptions provided in the original papers.

\section{Benchmark Evaluation Details}
\label{app:sec3}
\subsection{Implementation details}
\label{app:sec3.1}
All experiments use a unified evaluation framework. Key implementation details are as follows.

\noindent \textbf{Backbone MLLMs}: We evaluate three vision-language models via their official APIs: Qwen2.5-VL-3B-Instruct, Qwen2.5-VL-7B-Instruct~\cite{bai2025qwen25vltechnicalreport}, and GPT-4.1-Nano~\cite{openai2024gpt4technicalreport}. For each API call, we set the temperature to 0.1.

\noindent \textbf{Retriever}: Different embedding models are used for text‑only and multimodal retrieval. For text‑based methods, we adopt all-MiniLM-L6-v2~\cite{reimers-gurevych-2019-sentence} as the dense retriever. For multimodal methods, we use Alibaba-NLP/gme-Qwen2-VL-7B-Instruct~\cite{zhang2025gmeimprovinguniversalmultimodal}, which jointly encodes text and images. All retrievals are performed with a default top‑\(K=5\).

\noindent \textbf{Image Processing.} Our benchmark involves two types of images: (1) in-conversation images stored as part of the agent's memory, and (2) in-question images provided within the query during evaluation. For in-conversation images, storage format varies by memory method: text-based methods use GPT-4o~\cite{openai2024gpt4ocard} to generate image captions (limited to 256 tokens), replacing original images with textual descriptions stored alongside dialogue transcripts, while multimodal methods resize images to \(224 \times 224\) pixels and store them directly as visual tensors. During evaluation, text-based methods retrieve and reason over text-only memory entries (where images have been replaced by captions), whereas multimodal methods retrieve and utilize the original images alongside their associated text. For in-question images, all methods process them uniformly: images are resized to \(224 \times 224\) pixels and fed directly to the MLLM backbone together with the textual prompt.

\noindent \textbf{Computing Resources}: All experiments were conducted on a server with four NVIDIA A10 GPUs (24GB memory each). The total computational budget across all experiments is approximately \textbf{250--300 A10 GPU hours}.

\noindent \textbf{License and Terms of Use.} The embedding models (all-MiniLM-L6-v2~\cite{reimers-gurevych-2019-sentence}, Alibaba-NLP/gme-Qwen2-VL-7B-Instruct~\cite{zhang2025gmeimprovinguniversalmultimodal}) are downloaded from HuggingFace and used under their respective open-source licenses (Apache 2.0 and MIT). The Qwen2.5-VL series models (3B and 7B)~\cite{bai2025qwen25vltechnicalreport} are accessed via Alibaba Cloud's DashScope API and used in compliance with their terms of service. The GPT series models (GPT-4o~\cite{openai2024gpt4ocard}, GPT-4.1-Nano~\cite{openai2024gpt4technicalreport}, GPT-4o-mini) are accessed via OpenAI's API and used in compliance with OpenAI's terms of service. The H2HMem dataset is released under a CC BY 4.0 license for research purposes only.

\subsection{Evaluation Prompt Template}
\label{app:sec3.2}
All models receive a unified prompt template to ensure fair comparison. The template (Figure~\ref{fig:eval_prompt}) instructs the model to answer based solely on the provided conversation history (text and images) and to output the answer concisely. The temperature is fixed at 0.3 for all generations.

\subsection{Evaluation Metrics}
\label{app:sec3.3}
We evaluate all baseline methods from two perspectives: answer correctness and semantic alignment. All metrics are computed at the instance level and then averaged over the entire QA set.

\subsubsection{LLM-as-Judge}

\begin{figure}[htbp]
\centering
\begingroup
\setlength{\vskip}{0.3ex} 
\linespread{0.7}\selectfont
\begin{tcolorbox}[
    colback=gray!5,
    colframe=black,
    title=Judgment Prompt Template,
    width=1.0\textwidth,
    arc=3mm,
    boxrule=0.5pt,
    left=3mm,
    right=3mm,
    top=2.5mm,
    bottom=2.5mm,
    fonttitle=\bfseries
]

You are an impartial judge evaluating the memory capabilities of an AI assistant with the question-answering task.
Your task is to compare the Assistant's Answer against the Ground Truth and assign a score of 0, 0.25, 0.5, 0.75, or 1.

\noindent \textbf{Scoring Rubric:}

\noindent \textbf{Score 0 (Incorrect / Miss):}
\begin{itemize}
    \item The answer contradicts the Ground Truth.
    \item For Yes/No questions: The answer has the wrong polarity (e.g., says ``Yes'' when Ground Truth is ``No'').
    \item For Open-ended questions: The answer provides factually wrong information or hallucinations.
    \item The assistant fails to provide the required information.
\end{itemize}

\noindent \textbf{Score 0.25 (Poor / Tangential):}
\begin{itemize}
    \item The answer touches on the topic but misses the \textbf{core entity} or key value required.
    \item The answer contains a mix of minor correct details and \textbf{significant hallucinations} or wrong associations.
    \item The answer is excessively vague to the point of being useless (e.g., answering ``a dog'' instead of ``a golden retriever'').
\end{itemize}

\noindent \textbf{Score 0.5 (Partial / Vague):}
\begin{itemize}
    \item The answer is technically correct, but lacks confidence or is incomplete.
    \item The answer captures the \textbf{main entity or concept} correctly but misses a part of the required supporting details.
    \item For Yes/No questions: The polarity is correct, but the reasoning is flawed (if any), or the assistant is uncertain (e.g., ``I think it might be Yes'').
    \item For Open-ended questions: The answer is too general or misses key adjectives/details present in the Ground Truth.
\end{itemize}

\noindent \textbf{Score 0.75 (Good / Minor Imperfection):}
\begin{itemize}
    \item The answer is largely accurate and captures the core information confidently.
    \item It misses only \textbf{minor details} (e.g., specific adjectives or secondary details) that do not alter the main truth.
    \item The answer contains all the correct information but includes unnecessary ``fluff'' or slight conversational filler that reduces precision.
\end{itemize}

\noindent \textbf{Score 1 (Correct / Exact):}
\begin{itemize}
    \item The answer is accurate, precise, and confident.
    \item For Yes/No questions: The polarity matches the Ground Truth perfectly.
    \item For Open-ended questions: The answer contains \textbf{all} the core information and necessary details required by the Ground Truth without hallucinations.
\end{itemize}

\noindent \textbf{Input Data:}
\begin{verbatim}
Question: {{question}}
Ground Truth: {{ground_truth}}
Assistant Answer: {{model_output}}
\end{verbatim}

\noindent \textbf{Output Format:}\\
Output strictly in the following JSON format:\\
\verb|{"score": <0, 0.25, 0.5, 0.75, or 1>, "reasoning": "<short explanation>"}|

\end{tcolorbox}
\endgroup
\caption{Judgment prompt template for memory evaluation.}
\label{fig:judgment_prompt}
\end{figure}

Following prior work~\cite{zhang2026rethinkingevaluationretrievalaugmentedpersonalized}, we employ GPT-4o-mini as a zero-shot evaluator. The judge prompt (provided in Figure~\ref{fig:judgment_prompt}) asks the model to rate each response based on semantic equivalence to the ground truth, ignoring surface-level phrasing differences. 

To ensure the reliability of our primary evaluation metric, we validated the LLM-as-Judge approach against human judgments. A subset of 200 test instances was randomly sampled and independently annotated by two human evaluators. We measured the agreement between GPT-4o-mini's judgments and the human annotations using Cohen's $\kappa$, achieving a score of $\kappa = 0.84$.

\subsubsection{Lexical Metrics}
\label{app:sec3.4}
For tasks with deterministic ground-truth answers, we adopt four lexical overlap metrics.

\noindent \textbf{Precision (P)} measures the proportion of tokens in the predicted answer \(A_p\) that appear in the reference answer \(A_r\). Let \(T_p\) and \(T_r\) denote the multisets of tokens in \(A_p\) and \(A_r\), respectively. Precision is defined as:
\[
P = \frac{|T_p \cap T_r|}{|T_p|}
\]

\noindent \textbf{Recall (R)} measures the proportion of tokens in the reference answer \(A_r\) that are captured by the predicted answer \(A_p\):
\[
R = \frac{|T_p \cap T_r|}{|T_r|}
\]

\noindent \textbf{F1 Score} is the harmonic mean of precision and recall:
\[
F1 = \frac{2 \cdot P \cdot R}{P + R}
\]

\noindent \textbf{BLEU-1} measures unigram-level precision between the predicted answer \(A_p\) and the reference answer \(A_r\). Let \(\text{count}_{A_p}(w)\) denote the number of occurrences of unigram \(w\) in \(A_p\), and \(\text{count}_{A_r}(w)\) denote its occurrences in \(A_r\). The clipped count \(c(w)\) is defined as:
\[
c(w) = \min\bigl(\text{count}_{A_p}(w), \text{count}_{A_r}(w)\bigr)
\]
BLEU-1 is then computed as:
\[
\text{BLEU-1} = \frac{\sum_{w \in A_p} c(w)}{\sum_{w \in A_p} \text{count}_{A_p}(w)}
\]

\section{Additional Experimental Results}
\label{app:sec4}
\subsection{LLM-Judge Evaluation on Additional Backbones}
\label{app:sec4.1}

\begin{table}[htbp]
\centering
\caption{LLM-Judge performance of various methods on dyadic and multi-party conversations using Qwen2.5-3B-VL-Instruct~\cite{bai2025qwen25vltechnicalreport}. D = Dyadic, M = Multi-party, D\&M = Weighted average. {\color{blue}\textbf{*}} marks the higher value between D and M for the same method and metric. Bold numbers in D\&M rows indicate the highest value among all methods for LLM-Judge performance on the weighted average dataset. Light blue background highlights the D\&M data cells. }
\label{tab:dyadic_multiparty_combined_qwen_3B}
\newcommand{\dmcell}{\cellcolor{lightblue!80}}
\fontsize{7pt}{9pt}\selectfont
\setlength{\tabcolsep}{5pt}
\renewcommand{\arraystretch}{1.0}
\begin{tabular}{@{} l l l *{10}{l} @{}}
\toprule
\multirow{2}{*}{Category} & \multirow{2}{*}{Method} & \multirow{2}{*}{Dataset} & 
\multicolumn{3}{c}{Memory Recall} & \multicolumn{3}{c}{Memory Reasoning} & 
\multicolumn{3}{c}{Memory Application} & \multirow{2}{*}{Overall} \\
\cmidrule(lr){4-6} \cmidrule(lr){7-9} \cmidrule(lr){10-12}
 & & & UPR & CRR & KR & MCR & RET & TR & TTL & CD & AR & \\
\midrule
\multirow{9}{*}{Text-based} & \multirow{3}{*}{Full (Text)} & D & $0.2369$\textsuperscript{\textcolor{blue}{\textbf{*}}} & $0.2043$\textsuperscript{\textcolor{blue}{\textbf{*}}} & $0.3929$\textsuperscript{\textcolor{blue}{\textbf{*}}} & $0.2338$\textsuperscript{\textcolor{blue}{\textbf{*}}} & 0.2171 & $0.3846$\textsuperscript{\textcolor{blue}{\textbf{*}}} & 0.3463 & $0.2489$\textsuperscript{\textcolor{blue}{\textbf{*}}} & 0.9479 & 0.3284 \\
                           & & M & 0.3145 & 0.2600 & 0.3500 & 0.2000 & $0.3542$\textsuperscript{\textcolor{blue}{\textbf{*}}} & 0.3000 & $0.3900$\textsuperscript{\textcolor{blue}{\textbf{*}}} & 0.1200 & $0.9783$\textsuperscript{\textcolor{blue}{\textbf{*}}} & 0.3630\textsuperscript{\textcolor{blue}{\textbf{*}}} \\
                           & & D\&M & \dmcell 0.2439 & \dmcell 0.2080 & \dmcell 0.3880 & \dmcell 0.2314 & \dmcell 0.2269 & \dmcell 0.3752 & \dmcell 0.3514 & \dmcell 0.2358 & \dmcell 0.9507 & \dmcell 0.3313 \\
\cmidrule(lr){2-13}
                           & \multirow{3}{*}{NaiveRAG} & D & 0.5185 & 0.4207 & 0.4392 & 0.2855 & 0.3189 & 0.3500 & 0.5241 & $0.3341$\textsuperscript{\textcolor{blue}{\textbf{*}}} & $0.8957$\textsuperscript{\textcolor{blue}{\textbf{*}}} & 0.4667 \\
                           & & M & $0.6048$\textsuperscript{\textcolor{blue}{\textbf{*}}} & $0.6700$\textsuperscript{\textcolor{blue}{\textbf{*}}} & $0.4500$\textsuperscript{\textcolor{blue}{\textbf{*}}} & $0.3300$\textsuperscript{\textcolor{blue}{\textbf{*}}} & $0.4130$\textsuperscript{\textcolor{blue}{\textbf{*}}} & $0.4500$\textsuperscript{\textcolor{blue}{\textbf{*}}} & $0.5900$\textsuperscript{\textcolor{blue}{\textbf{*}}} & 0.2000 & 0.8229 & $0.5173$\textsuperscript{\textcolor{blue}{\textbf{*}}} \\
                           & & D\&M & \dmcell 0.5263 & \dmcell 0.4370 & \dmcell 0.4404 & \dmcell 0.2887 & \dmcell 0.3257 & \dmcell 0.3611 & \dmcell 0.5318 & \dmcell 0.3204 & \dmcell 0.8890 & \dmcell 0.4607 \\
\cmidrule(lr){2-13}
                           & \multirow{3}{*}{A-Mem~\cite{xu2026amem}} & D & $0.6786$\textsuperscript{\textcolor{blue}{\textbf{*}}} & 0.5453 & $0.4872$\textsuperscript{\textcolor{blue}{\textbf{*}}} & 0.3735 & 0.3349 & $0.4125$\textsuperscript{\textcolor{blue}{\textbf{*}}} & 0.6053 & $0.3664$\textsuperscript{\textcolor{blue}{\textbf{*}}} & 0.9258 & 0.5273 \\
                           & & M & 0.6774 & $0.6500$\textsuperscript{\textcolor{blue}{\textbf{*}}} & 0.3000 & 0.3200 & $0.3854$\textsuperscript{\textcolor{blue}{\textbf{*}}} & 0.4000 & $0.6300$\textsuperscript{\textcolor{blue}{\textbf{*}}} & 0.2400 & $1.0000$\textsuperscript{\textcolor{blue}{\textbf{*}}} & 0.5465\textsuperscript{\textcolor{blue}{\textbf{*}}} \\
                           & & D\&M & \dmcell \textbf{0.6785} & \dmcell \textbf{0.5521} & \dmcell 0.4659 & \dmcell 0.3697 & \dmcell 0.3385 & \dmcell \textbf{0.4111} & \dmcell \textbf{0.6082} & \dmcell 0.3536 & \dmcell 0.9326 & \dmcell \textbf{0.5292} \\
\midrule
\multirow{9}{*}{Multi-modal} & \multirow{3}{*}{Full (MM)} & D & $0.3493$\textsuperscript{\textcolor{blue}{\textbf{*}}} & $0.2783$\textsuperscript{\textcolor{blue}{\textbf{*}}} & $0.5128$\textsuperscript{\textcolor{blue}{\textbf{*}}} & $0.2384$\textsuperscript{\textcolor{blue}{\textbf{*}}} & 0.2913 & 0.3654 & 0.4535 & $0.2283$\textsuperscript{\textcolor{blue}{\textbf{*}}} & $0.9516$\textsuperscript{\textcolor{blue}{\textbf{*}}} & 0.3829\textsuperscript{\textcolor{blue}{\textbf{*}}} \\
                           & & M & 0.3250 & 0.2188 & 0.3500 & 0.2188 & $0.4062$\textsuperscript{\textcolor{blue}{\textbf{*}}} & $0.4000$\textsuperscript{\textcolor{blue}{\textbf{*}}} & $0.5300$\textsuperscript{\textcolor{blue}{\textbf{*}}} & 0.0600 & 0.9375 & 0.3817 \\
                           & & D\&M & \dmcell 0.3471 & \dmcell 0.2744 & \dmcell 0.4943 & \dmcell 0.2370 & \dmcell 0.2995 & \dmcell 0.3692 & \dmcell 0.4624 & \dmcell 0.2112 & \dmcell 0.9503 & \dmcell 0.3827 \\
\cmidrule(lr){2-13}
                           & \multirow{3}{*}{MuRAG~\cite{chen-etal-2022-murag}} & D & 0.6115 & 0.5348 & $0.4539$\textsuperscript{\textcolor{blue}{\textbf{*}}} & 0.3927 & 0.3604 & $0.4062$\textsuperscript{\textcolor{blue}{\textbf{*}}} & 0.5865 & $0.3608$\textsuperscript{\textcolor{blue}{\textbf{*}}} & 0.9128 & 0.5236 \\
                           & & M & $0.6210$\textsuperscript{\textcolor{blue}{\textbf{*}}} & $0.6500$\textsuperscript{\textcolor{blue}{\textbf{*}}} & 0.4500 & $0.4600$\textsuperscript{\textcolor{blue}{\textbf{*}}} & $0.4062$\textsuperscript{\textcolor{blue}{\textbf{*}}} & 0.2000 & $0.6300$\textsuperscript{\textcolor{blue}{\textbf{*}}} & 0.1600 & $1.0000$\textsuperscript{\textcolor{blue}{\textbf{*}}} & 0.5489\textsuperscript{\textcolor{blue}{\textbf{*}}} \\
                           & & D\&M & \dmcell 0.6124 & \dmcell 0.5423 & \dmcell 0.4535 & \dmcell 0.3974 & \dmcell 0.3637 & \dmcell 0.3833 & \dmcell 0.5916 & \dmcell 0.3405 & \dmcell 0.9208 & \dmcell 0.5257 \\
\cmidrule(lr){2-13}
                           & \multirow{3}{*}{NGM~\cite{fisher2025neural}} & D & 0.4738 & $0.4351$\textsuperscript{\textcolor{blue}{\textbf{*}}} & $0.5128$\textsuperscript{\textcolor{blue}{\textbf{*}}} & 0.3182 & 0.3213 & $0.3250$\textsuperscript{\textcolor{blue}{\textbf{*}}} & 0.5548 & $0.4206$\textsuperscript{\textcolor{blue}{\textbf{*}}} & 0.9645 & 0.4671 \\
                           & & M & $0.4758$\textsuperscript{\textcolor{blue}{\textbf{*}}} & 0.4200 & 0.4000 & 0.3100 & $0.3750$\textsuperscript{\textcolor{blue}{\textbf{*}}} & 0.3000 & $0.7300$\textsuperscript{\textcolor{blue}{\textbf{*}}} & 0.1200 & $1.0000$\textsuperscript{\textcolor{blue}{\textbf{*}}} & 0.4802\textsuperscript{\textcolor{blue}{\textbf{*}}} \\
                           & & D\&M & \dmcell 0.4740 & \dmcell 0.4341 & \dmcell 0.5000 & \dmcell 0.3176 & \dmcell 0.3251 & \dmcell 0.3222 & \dmcell 0.5752 & \dmcell 0.3902 & \dmcell 0.9678 & \dmcell 0.4682 \\
\bottomrule
\end{tabular}
\end{table}

\begin{table}[htbp]
\centering
\caption{LLM-Judge performance of various methods on dyadic and multi-party conversations using Qwen2.5-VL-7B-Instruct~\cite{bai2025qwen25vltechnicalreport}. D = Dyadic, M = Multi-party, D\&M = Weighted average. {\color{blue}\textbf{*}} marks the higher value between D and M for the same method and metric. Bold numbers in D\&M rows indicate the highest value among all methods for LLM-Judge performance on the weighted average dataset. Light blue background highlights the D\&M data cells.}
\label{tab:dyadic_multiparty_combined_qwen_7B}
\fontsize{7pt}{9pt}\selectfont
\setlength{\tabcolsep}{5pt}
\newcommand{\dmcell}{\cellcolor{lightblue!80}}
\renewcommand{\arraystretch}{1.0}
\begin{tabular}{@{} l l l *{10}{l} @{}}
\toprule
\multirow{2}{*}{Category} & \multirow{2}{*}{Method} & \multirow{2}{*}{Dataset} & 
\multicolumn{3}{c}{Memory Recall} & \multicolumn{3}{c}{Memory Reasoning} & 
\multicolumn{3}{c}{Memory Application} & \multirow{2}{*}{Overall} \\
\cmidrule(lr){4-6} \cmidrule(lr){7-9} \cmidrule(lr){10-12}
 & & & UPR & CRR & KR & MCR & RET & TR & TTL & CD & AR & \\
\midrule
\multirow{9}{*}{Text-based} & \multirow{3}{*}{Full (Text)} & D & $0.2667$\textsuperscript{\textcolor{blue}{\textbf{*}}} & 0.2352 & $0.3716$\textsuperscript{\textcolor{blue}{\textbf{*}}} & $0.2389$\textsuperscript{\textcolor{blue}{\textbf{*}}} & $0.2632$\textsuperscript{\textcolor{blue}{\textbf{*}}} & $0.4392$\textsuperscript{\textcolor{blue}{\textbf{*}}} & 0.3878 & $0.2661$\textsuperscript{\textcolor{blue}{\textbf{*}}} & 0.8817 & 0.3480\textsuperscript{\textcolor{blue}{\textbf{*}}} \\
                           & & M & 0.1897 & $0.3646$\textsuperscript{\textcolor{blue}{\textbf{*}}} & 0.2500 & 0.1800 & 0.2500 & 0.2000 & $0.3900$\textsuperscript{\textcolor{blue}{\textbf{*}}} & 0.1562 & $0.9457$\textsuperscript{\textcolor{blue}{\textbf{*}}} & 0.3388 \\
                           & & D\&M & \dmcell 0.2630 & \dmcell 0.2408 & \dmcell 0.3650 & \dmcell 0.2356 & \dmcell 0.2625 & \dmcell 0.4227 & \dmcell 0.3880 & \dmcell 0.2589 & \dmcell 0.8861 & \dmcell 0.3474 \\
\cmidrule(lr){2-13}
                           & \multirow{3}{*}{NaiveRAG} & D & 0.4918 & 0.4096 & $0.5312$\textsuperscript{\textcolor{blue}{\textbf{*}}} & 0.1870 & 0.2795 & 0.2143 & 0.5156 & $0.2066$\textsuperscript{\textcolor{blue}{\textbf{*}}} & 0.8643 & 0.4127 \\
                           & & M & $0.5104$\textsuperscript{\textcolor{blue}{\textbf{*}}} & 0.3889 & 0.1000 & $0.2083$\textsuperscript{\textcolor{blue}{\textbf{*}}} & 0.1719 & $0.2500$\textsuperscript{\textcolor{blue}{\textbf{*}}} & $0.6806$\textsuperscript{\textcolor{blue}{\textbf{*}}} & 0.1389 & $0.9444$\textsuperscript{\textcolor{blue}{\textbf{*}}} & 0.4232\textsuperscript{\textcolor{blue}{\textbf{*}}} \\
                           & & D\&M & \dmcell 0.4937 & \dmcell 0.4067 & \dmcell 0.4741 & \dmcell 0.1891 & \dmcell 0.2685 & \dmcell 0.2183 & \dmcell 0.5372 & \dmcell 0.1990 & \dmcell 0.8733 & \dmcell 0.4135 \\
\cmidrule(lr){2-13}
                           & \multirow{3}{*}{A-Mem~\cite{xu2026amem}} & D & 0.6330 & 0.5524 & $0.5577$\textsuperscript{\textcolor{blue}{\textbf{*}}} & 0.4126 & 0.3955 & $0.5000$\textsuperscript{\textcolor{blue}{\textbf{*}}} & 0.6144 & $0.3909$\textsuperscript{\textcolor{blue}{\textbf{*}}} & 0.8860 & 0.5429 \\
                           & & M & 0.6048 & $0.6700$\textsuperscript{\textcolor{blue}{\textbf{*}}} & 0.3500 & $0.5100$\textsuperscript{\textcolor{blue}{\textbf{*}}} & $0.4167$\textsuperscript{\textcolor{blue}{\textbf{*}}} & 0.4000 & 0.5833 & 0.0800 & 0.8854 & 0.5279 \\
                           & & D\&M & \dmcell \textbf{0.6306} & \dmcell \textbf{0.5585} & \dmcell 0.5405 & \dmcell \textbf{0.4201} & \dmcell \textbf{0.3977} & \dmcell \textbf{0.4914} & \dmcell \textbf{0.6113} & \dmcell 0.3656 & \dmcell 0.8860 & \dmcell \textbf{0.5415} \\
\midrule
\multirow{9}{*}{Multi-modal} & \multirow{3}{*}{Full (MM)} & D & $0.3681$\textsuperscript{\textcolor{blue}{\textbf{*}}} & $0.2514$\textsuperscript{\textcolor{blue}{\textbf{*}}} & $0.5203$\textsuperscript{\textcolor{blue}{\textbf{*}}} & $0.2640$\textsuperscript{\textcolor{blue}{\textbf{*}}} & 0.2782 & $0.3875$\textsuperscript{\textcolor{blue}{\textbf{*}}} & 0.3889 & $0.1714$\textsuperscript{\textcolor{blue}{\textbf{*}}} & 0.7820 & 0.3512\textsuperscript{\textcolor{blue}{\textbf{*}}} \\
                           & & M & 0.3190 & 0.2083 & 0.4500 & 0.2292 & $0.2727$\textsuperscript{\textcolor{blue}{\textbf{*}}} & 0.3500 & 0.3854 & 0.1042 & $0.8478$\textsuperscript{\textcolor{blue}{\textbf{*}}} & 0.3389 \\
                           & & D\&M & \dmcell 0.3651 & \dmcell 0.2488 & \dmcell 0.5148 & \dmcell 0.2617 & \dmcell 0.2779 & \dmcell 0.3844 & \dmcell 0.3884 & \dmcell 0.1660 & \dmcell 0.7880 & \dmcell 0.3503 \\
\cmidrule(lr){2-13}
                           & \multirow{3}{*}{MuRAG~\cite{chen-etal-2022-murag}} & D & 0.5924 & 0.5271 & $0.5705$\textsuperscript{\textcolor{blue}{\textbf{*}}} & 0.3843 & 0.3640 & $0.4437$\textsuperscript{\textcolor{blue}{\textbf{*}}} & 0.5726 & $0.3553$\textsuperscript{\textcolor{blue}{\textbf{*}}} & 0.7998 & 0.5053 \\
                           & & M & 0.5645 & $0.6700$\textsuperscript{\textcolor{blue}{\textbf{*}}} & 0.4500 & $0.4700$\textsuperscript{\textcolor{blue}{\textbf{*}}} & $0.3750$\textsuperscript{\textcolor{blue}{\textbf{*}}} & 0.2000 & $0.6400$\textsuperscript{\textcolor{blue}{\textbf{*}}} & 0.1250 & $1.0000$\textsuperscript{\textcolor{blue}{\textbf{*}}} & 0.5361\textsuperscript{\textcolor{blue}{\textbf{*}}} \\
                           & & D\&M & \dmcell 0.5899 & \dmcell 0.5356 & \dmcell \textbf{0.5579} & \dmcell 0.3904 & \dmcell 0.3652 & \dmcell 0.4220 & \dmcell 0.5818 & \dmcell 0.3342 & \dmcell 0.8196 & \dmcell 0.5079 \\
\cmidrule(lr){2-13}
                           & \multirow{3}{*}{NGM~\cite{fisher2025neural}} & D & 0.4976 & $0.4630$\textsuperscript{\textcolor{blue}{\textbf{*}}} & $0.4936$\textsuperscript{\textcolor{blue}{\textbf{*}}} & 0.3182 & $0.3798$\textsuperscript{\textcolor{blue}{\textbf{*}}} & $0.4000$\textsuperscript{\textcolor{blue}{\textbf{*}}} & 0.6000 & $0.3779$\textsuperscript{\textcolor{blue}{\textbf{*}}} & 0.8721 & 0.4782 \\
                           & & M & 0.4758 & 0.4200 & 0.4000 & 0.3100 & 0.3750 & 0.3000 & $0.7300$\textsuperscript{\textcolor{blue}{\textbf{*}}} & 0.1200 & $1.0000$\textsuperscript{\textcolor{blue}{\textbf{*}}} & 0.4802\textsuperscript{\textcolor{blue}{\textbf{*}}} \\
                           & & D\&M & \dmcell 0.4956 & \dmcell 0.4591 & \dmcell 0.4864 & \dmcell 0.3174 & \dmcell 0.3794 & \dmcell 0.3927 & \dmcell 0.6160 & \dmcell \textbf{0.3567} & \dmcell \textbf{0.8847} & \dmcell 0.4783 \\
\bottomrule
\end{tabular}
\end{table}

In the main paper, we report LLM-Judge results using GPT-4.1-Nano~\cite{openai2024gpt4technicalreport}. Here, we extend the evaluation to two additional vision-language backbones: Qwen2.5-VL-3B-Instruct and Qwen2.5-VL-7B-Instruct~\cite{bai2025qwen25vltechnicalreport}. Table~\ref{tab:dyadic_multiparty_combined_qwen_3B} and Table~\ref{tab:dyadic_multiparty_combined_qwen_7B} present the results.

We highlight three key observations that directly reflect the unique characteristics of our multimodal human–human interaction benchmark:

\noindent \textbf{(i) Overall performance is consistently low across backbones.}
Across three backbones, the absolute scores of all methods remain below 0.6 (on a 0–1 scale), with the best overall weighted average reaching only 0.5757 (A-Mem~\cite{xu2026amem} on GPT-4.1-Nano~\cite{openai2024gpt4technicalreport}). Even with strong memory mechanisms, models struggle to exceed this ceiling. This uniformly limited performance underscores the fundamental difficulty of our benchmark: agents must integrate information scattered across multiple participants, modalities (text and images), and sessions – a far cry from traditional single‑turn or dyadic QA.

\noindent \textbf{(ii) Limited benefits from model scaling for cross-modal and multi-party reasoning.}
Increasing parameter count from 3B to 7B yields only moderate gains, and tasks central to our benchmark — Cross-modal Related Retrieval (CRR), Multimodal Causal Reasoning (MCR), and Conflict Detection (CD) — show minimal improvement. This suggests that larger models alone cannot resolve the inherent difficulties of aligning information across participants and modalities, or of tracking evolving references across multiple sessions.

\noindent \textbf{(iii) Persistent bottlenecks in reasoning over multimodal, human–human memory.}
Across all backbones and methods, tasks that require structured reasoning over distributed evidence — especially Multimodal Causal Reasoning (MCR), Reference \& Evolution Tracking (RET), and Conflict Detection (CD) in multi‑party dialogues — remain far below recall‑oriented tasks. This gap confirms that current models lack robust mechanisms for maintaining coherent memory across time, speakers, and modalities, which is the central challenge posed by our human–human interaction benchmark.

These results reinforce that current systems struggle with organizing and utilizing fragmented memory, especially under the multimodal, human--human interaction captured by our benchmark. The consistency of these findings across different backbone models demonstrates that the challenges identified—cross-source alignment, structured reasoning over distributed evidence, and robustness to incomplete retrieval—are fundamental rather than artifacts of a particular model choice. Importantly, the persistence of these limitations across both proprietary and open-source backbones highlights the necessity of benchmarking conversational memory in realistic human–human interaction settings, moving beyond simplified dyadic, single-modality evaluations.

\subsection{Lexical-Level Evaluation on Additional Backbones}
\label{app:sec4.2}
\begin{table}[htbp]
\centering
\caption{Weighted average (D\&M) performance of different methods (Qwen2.5-3B-VL-Instruct~\cite{bai2025qwen25vltechnicalreport}, top-5 retrieval) across categories. Metrics: P=Precision, R=Recall, F1=F1-score, B=BLEU-1. Bold values indicate the best performance among the six methods within each metric column for the given category.}
\label{tab:lexical_metrics_Qwen2.5-3B-VL-Instruct}
\fontsize{7pt}{8.5pt}\selectfont
\setlength{\tabcolsep}{6pt}
\renewcommand{\arraystretch}{1.0}
\begin{tabular}{@{} l l l *{10}{c} @{}}
\toprule
\multirow{2}{*}{Category} & \multirow{2}{*}{Method} & \multirow{2}{*}{Metrics} & 
\multicolumn{3}{c}{Memory Recall} & \multicolumn{3}{c}{Memory Reasoning} & 
\multicolumn{3}{c}{Memory Application} & \multirow{2}{*}{Overall} \\
\cmidrule(lr){4-6} \cmidrule(lr){7-9} \cmidrule(lr){10-12}
 & & & UPR & CRR & KR & MCR & RET & TR & TTL & CD & AR & \\
\midrule
\multirow{12}{*}{Text-based} 
 & \multirow{4}{*}{Full (Text)} 
   & P & 0.1274 & 0.0830 & 0.3095 & 0.0950 & 0.1486 & 0.3775 & 0.1334 & 0.0448 & \textbf{0.9433} & 0.2293 \\
   & & R & 0.2236 & 0.1922 & 0.2966 & 0.2074 & 0.1788 & 0.4185 & 0.1977 & 0.0360 & \textbf{0.9361} & 0.2548 \\
   & & F1& 0.1428 & 0.1008 & 0.2862 & 0.1113 & 0.1348 & 0.3187 & 0.1337 & 0.0372 & \textbf{0.9344} & 0.2295 \\
   & & B & 0.1049 & 0.0736 & 0.2159 & 0.0807 & 0.0887 & \textbf{0.2038} & 0.1096 & 0.0340 & \textbf{0.9322} & 0.2206 \\
\cmidrule(lr){2-13}
 & \multirow{4}{*}{NaiveRAG} 
   & P & \textbf{0.2148} & \textbf{0.1686} & \textbf{0.4127} & \textbf{0.1164} & \textbf{0.1550} & \textbf{0.4592} & \textbf{0.1693} & \textbf{0.1932} & 0.8733 & \textbf{0.2992} \\
   & & R & 0.3776 & 0.3079 & 0.3066 & 0.1959 & 0.2368 & 0.4288 & 0.3582 & \textbf{0.1694} & 0.8670 & 0.2957 \\
   & & F1& \textbf{0.2390} & \textbf{0.1888} & 0.3383 & 0.1283 & 0.1649 & \textbf{0.3596} & \textbf{0.1960} & \textbf{0.1734} & 0.8662 & \textbf{0.2894} \\
   & & B & \textbf{0.1888} & \textbf{0.1474} & 0.2330 & \textbf{0.0976} & \textbf{0.1220} & 0.2003 & \textbf{0.1481} & \textbf{0.1663} & 0.8640 & \textbf{0.2743} \\
\cmidrule(lr){2-13}
 & \multirow{4}{*}{A-Mem~\cite{xu2026amem}} 
   & P & 0.1599 & 0.1120 & 0.3608 & 0.1027 & 0.1342 & 0.2264 & 0.1071 & 0.0987 & 0.8959 & 0.2051 \\
   & & R & \textbf{0.4590} & \textbf{0.4143} & \textbf{0.4391} & \textbf{0.3279} & \textbf{0.3326} & \textbf{0.4899} & \textbf{0.4297} & 0.0710 & 0.9120 & \textbf{0.4160} \\
   & & F1& 0.2134 & 0.1622 & \textbf{0.3832} & \textbf{0.1412} & \textbf{0.1694} & 0.2698 & 0.1568 & 0.0733 & 0.8915 & 0.2257 \\
   & & B & 0.1496 & 0.1000 & \textbf{0.3241} & 0.0931 & 0.1195 & 0.1915 & 0.1026 & 0.0016 & 0.8848 & 0.2038 \\
\midrule
\multirow{12}{*}{Multi-modal} 
 & \multirow{4}{*}{Full (MM)} 
   & P & 0.1681 & 0.0891 & 0.3617 & 0.1077 & 0.1481 & \textbf{0.5761} & 0.1261 & 0.0492 & \textbf{0.9390} & 0.2196 \\
   & & R & 0.2329 & 0.1516 & 0.3822 & 0.1610 & 0.2295 & 0.4177 & 0.2593 & 0.0434 & \textbf{0.9323} & 0.2967 \\
   & & F1& 0.1703 & 0.0979 & 0.3540 & 0.1160 & 0.1493 & 0.4057 & 0.1346 & 0.0427 & \textbf{0.9303} & 0.2194 \\
   & & B & 0.1394 & 0.0774 & 0.2809 & 0.0929 & 0.1042 & 0.2239 & 0.0972 & \textbf{0.0418} & \textbf{0.9286} & 0.2086 \\
\cmidrule(lr){2-13}
 & \multirow{4}{*}{MuRAG~\cite{chen-etal-2022-murag}} 
   & P & 0.1959 & 0.1272 & 0.3653 & 0.1365 & 0.1449 & 0.3337 & 0.1425 & \textbf{0.0729} & 0.9323 & 0.2424 \\
   & & R & \textbf{0.4253} & \textbf{0.3320} & \textbf{0.3862} & \textbf{0.2932} & \textbf{0.3220} & \textbf{0.5310} & 0.3537 & \textbf{0.0539} & 0.9241 & \textbf{0.3371} \\
   & & F1& 0.2349 & 0.1661 & \textbf{0.3653} & 0.1693 & 0.1749 & 0.3426 & 0.1735 & \textbf{0.0556} & 0.9219 & 0.2666 \\
   & & B & 0.1528 & 0.1109 & \textbf{0.3081} & 0.1229 & 0.1252 & 0.2357 & 0.1061 & 0.0000 & 0.9149 & 0.2369 \\
\cmidrule(lr){2-13}
 & \multirow{4}{*}{NGM~\cite{fisher2025neural}} 
   & P & \textbf{0.3179} & \textbf{0.1627} & \textbf{0.3839} & \textbf{0.1579} & \textbf{0.1805} & 0.5122 & \textbf{0.2573} & 0.0490 & 0.9365 & \textbf{0.2866} \\
   & & R & 0.3763 & 0.2830 & 0.3196 & 0.2575 & 0.2556 & 0.4434 & \textbf{0.3672} & 0.0366 & 0.9311 & 0.3244 \\
   & & F1& \textbf{0.3107} & \textbf{0.1845} & 0.3345 & \textbf{0.1736} & \textbf{0.1867} & \textbf{0.4068} & \textbf{0.2738} & 0.0369 & 0.9284 & \textbf{0.2792} \\
   & & B & \textbf{0.2011} & \textbf{0.1213} & 0.2489 & \textbf{0.1267} & \textbf{0.1310} & \textbf{0.2473} & \textbf{0.1768} & 0.0000 & 0.9249 & \textbf{0.2607} \\
\bottomrule
\end{tabular}
\end{table}

Tables~\ref{tab:lexical_metrics_Qwen2.5-3B-VL-Instruct} and \ref{tab:lexical_metrics_Qwen2.5-7B-VL-Instruct} report additional lexical-level results on Qwen2.5-VL-3B-Instruct and Qwen2.5-VL-7B-Instruct~\cite{bai2025qwen25vltechnicalreport}, complementing the main findings presented in the paper using GPT-4.1-Nano~\cite{openai2024gpt4technicalreport}.

\noindent \textbf{Findings are consistent across backbone models.}
Overall, the trends observed in the main paper remain stable across both smaller and larger Qwen backbones. In particular, lexical metrics remain uniformly low (generally below 0.4), reinforcing the conclusion that our benchmark poses a fundamentally challenging setting where exact lexical overlap is difficult due to distributed, multi-session, and multimodal information sources. This consistency suggests that the difficulty is not tied to a specific backbone, but is intrinsic to the task.

\begin{table}[htbp]
\centering
\caption{Weighted average (D\&M) performance of different methods (Qwen2.5-7B-VL-Instruct~\cite{bai2025qwen25vltechnicalreport}, top-5 retrieval) across categories. Metrics: P=Precision, R=Recall, F1=F1-score, B=BLEU-1. Bold values indicate the best performance among the six methods within each metric column for the given category.}
\label{tab:lexical_metrics_Qwen2.5-7B-VL-Instruct}
\fontsize{7pt}{8.5pt}\selectfont
\setlength{\tabcolsep}{6pt}
\renewcommand{\arraystretch}{1.0}
\begin{tabular}{@{} l l l *{10}{c} @{}}  
\toprule
\multirow{2}{*}{Category} & \multirow{2}{*}{Method} & \multirow{2}{*}{Metrics} & 
\multicolumn{3}{c}{Memory Recall} & \multicolumn{3}{c}{Memory Reasoning} & 
\multicolumn{3}{c}{Memory Application} & \multirow{2}{*}{Overall} \\
\cmidrule(lr){4-6} \cmidrule(lr){7-9} \cmidrule(lr){10-12}
 & & & UPR & CRR & KR & MCR & RET & TR & TTL & CD & AR &\\
\midrule
\multirow{12}{*}{Text-based} 
 & \multirow{4}{*}{Full (Text)} 
   & P & 0.0784 & 0.0599 & 0.1913 & 0.0712 & 0.0764 & 0.3516 & 0.0826 & 0.0528 & \textbf{0.8654} & 0.1745 \\
   & & R & 0.2014 & 0.2270 & 0.3362 & 0.2186 & 0.2332 & 0.4880 & 0.2409 & 0.0393 & \textbf{0.8574} & 0.2871 \\
   & & F1& 0.0984 & 0.0822 & 0.2358 & 0.0945 & 0.1014 & 0.3273 & 0.1023 & 0.0410 & \textbf{0.8575} & 0.1888 \\
   & & B & 0.0699 & 0.0529 & 0.1988 & 0.0668 & 0.0724 & 0.2279 & 0.0709 & 0.0372 & \textbf{0.8535} & 0.1641 \\
\cmidrule(lr){2-13}
 & \multirow{4}{*}{NaiveRAG} 
   & P & \textbf{0.2876} & \textbf{0.1741} & \textbf{0.2779} & 0.0837 & \textbf{0.1283} & 0.1629 & \textbf{0.1536} & \textbf{0.0716} & 0.8571 & 0.2560 \\
   & & R & 0.2999 & 0.1952 & 0.1328 & 0.0868 & 0.1151 & 0.2654 & 0.2208 & \textbf{0.0599} & 0.8515 & 0.2819 \\
   & & F1& \textbf{0.2624} & \textbf{0.1663} & 0.1666 & 0.0731 & 0.0990 & 0.1288 & \textbf{0.1629} & \textbf{0.0607} & 0.8500 & 0.2569 \\
   & & B & \textbf{0.2177} & \textbf{0.1325} & 0.0605 & 0.0533 & 0.0558 & 0.0752 & \textbf{0.1298} & \textbf{0.0573} & 0.8476 & 0.2415 \\
\cmidrule(lr){2-13}
 & \multirow{4}{*}{A-Mem~\cite{xu2026amem}} 
   & P & 0.0739 & 0.0500 & 0.1962 & 0.0627 & 0.0624 & 0.0943 & 0.0506 & 0.0545 & 0.5624 & 0.1894 \\
   & & R & \textbf{0.4381} & \textbf{0.4206} & \textbf{0.3834} & \textbf{0.3074} & \textbf{0.3397} & 0.5892 & \textbf{0.3428} & 0.0427 & 0.8198 & \textbf{0.4107} \\
   & & F1& 0.1185 & 0.0822 & 0.2519 & 0.0939 & 0.0982 & 0.1465 & 0.0826 & 0.0425 & 0.5707 & 0.1943 \\
   & & B & 0.0972 & 0.0385 & 0.2024 & 0.0586 & 0.0680 & 0.0955 & 0.0569 & 0.0010 & 0.5555 & 0.1702 \\
\midrule
\multirow{12}{*}{Multi-modal} 
 & \multirow{4}{*}{Full (MM)} 
   & P & 0.0971 & 0.0470 & 0.1770 & 0.0494 & 0.0524 & 0.2606 & 0.0477 & 0.0535 & 0.7640 & 0.2047 \\
   & & R & 0.2361 & 0.1862 & 0.3422 & 0.1761 & 0.2606 & 0.5056 & 0.2412 & 0.0411 & 0.7578 & 0.2936 \\
   & & F1& 0.1171 & 0.0633 & 0.2244 & 0.0683 & 0.0763 & 0.2210 & 0.0677 & 0.0426 & 0.7546 & 0.2070 \\
   & & B & 0.0894 & 0.0420 & 0.1810 & 0.0474 & 0.0529 & 0.1208 & 0.0430 & 0.0394 & 0.7431 & 0.1935 \\
\cmidrule(lr){2-13}
 & \multirow{4}{*}{MuRAG~\cite{chen-etal-2022-murag}} 
   & P & 0.1247 & 0.0716 & 0.2083 & 0.0929 & 0.0745 & 0.1945 & 0.0756 & 0.0554 & 0.7862 & 0.2286 \\
   & & R & 0.3999 & 0.3402 & 0.3413 & 0.2774 & 0.3070 & \textbf{0.5959} & 0.3282 & 0.0468 & 0.8220 & 0.3330 \\
   & & F1& 0.1692 & 0.1032 & 0.2518 & 0.1241 & 0.1069 & 0.2249 & 0.1046 & 0.0483 & 0.7805 & 0.2354 \\
   & & B & 0.1279 & 0.0607 & 0.2055 & 0.0872 & 0.0781 & 0.1524 & 0.0617 & 0.0020 & 0.7664 & 0.2089 \\
\cmidrule(lr){2-13}
 & \multirow{4}{*}{NGM~\cite{fisher2025neural}} 
   & P & 0.1925 & 0.0884 & 0.2652 & \textbf{0.1070} & 0.0981 & \textbf{0.4642} & 0.1246 & 0.0605 & 0.8435 & \textbf{0.2732} \\
   & & R & 0.3505 & 0.2925 & 0.3155 & 0.2393 & 0.2755 & 0.5359 & 0.3089 & 0.0487 & 0.8461 & 0.3213 \\
   & & F1& 0.2185 & 0.1149 & \textbf{0.2784} & \textbf{0.1372} & \textbf{0.1262} & \textbf{0.4224} & 0.1588 & 0.0497 & 0.8391 & \textbf{0.2678} \\
   & & B & 0.1478 & 0.0760 & \textbf{0.2223} & \textbf{0.0978} & \textbf{0.0989} & \textbf{0.2751} & 0.1001 & 0.0000 & 0.8276 & \textbf{0.2427} \\
\bottomrule
\end{tabular}
\end{table}

\noindent \textbf{External memory consistently improves recall.}
Across both Qwen models, methods with external memory (e.g., NaiveRAG, MuRAG~\cite{chen-etal-2022-murag}, A-Mem~\cite{xu2026amem}, NGM~\cite{fisher2025neural}) substantially outperform full-context baselines in recall, mirroring the behavior observed with GPT-4.1-Nano~\cite{openai2024gpt4technicalreport}. For example, A-Mem~\cite{xu2026amem} achieves the highest recall across most settings (e.g., 0.4160 on Qwen2.5-VL-3B-Instruct~\cite{bai2025qwen25vltechnicalreport} and 0.4107 on Qwen2.5-VL-3B-Instruct~\cite{bai2025qwen25vltechnicalreport}), indicating that structured memory access is critical for recovering dispersed information. However, precision remains low across all methods, confirming that retrieved evidence often contains noise.

\noindent \textbf{Cross-modal retrieval remains a bottleneck.}
The gap between unimodal recall (UPR) and cross-modal retrieval (CRR) persists across both Qwen backbones. Even multimodal retrieval methods such as MuRAG~\cite{chen-etal-2022-murag} and NGM~\cite{fisher2025neural} fail to close this gap, highlighting the continued difficulty of aligning textual queries with visual content in human–human interaction settings.

\noindent \textbf{Reasoning-intensive tasks show extremely low lexical overlap.}
Tasks such as MCR, RET, and especially CD continue to exhibit near-zero BLEU-1 scores across both models. This further confirms that lexical metrics are poorly suited for evaluating open-ended reasoning and decision-making tasks, and that models struggle to produce lexically aligned outputs even when reasoning is partially correct.

\noindent \textbf{Model scaling provides limited gains.}
Comparing Qwen2.5-VL-3B-Instruct and Qwen2.5-VL-7B-Instruct~\cite{bai2025qwen25vltechnicalreport}, improvements from scaling are modest and inconsistent. While the 7B model shows slight gains in some recall metrics, it does not fundamentally change the performance landscape. This observation aligns with the main paper: increasing model capacity alone is insufficient to address challenges in memory retrieval, cross-modal grounding, and multi-session reasoning.

\subsection{Retriever Analysis}
Table~\ref{tab:topk_results} reports the performance under different top-$K$ retrieval settings. We observe that increasing $K$ generally improves performance for most methods, suggesting that retrieving more candidate memories helps alleviate missing relevant information. However, the gain is not monotonic. For example, A-Mem~\cite{xu2026amem} and NGM~\cite{fisher2025neural} both achieve their best performance at $K=15$, after which performance slightly declines. This indicates that introducing excessive retrieved content may introduce noise and negatively affect downstream reasoning. Different methods also exhibit varying sensitivity to $K$. NaiveRAG shows a relatively steady improvement as $K$ increases, implying that its simple retrieval-and-generate pipeline primarily benefits from increased recall. In contrast, MuRAG~\cite{chen-etal-2022-murag} reaches its peak at $K=10$ and then degrades, suggesting that multimodal retrieval may be more susceptible to noise accumulation when irrelevant information is introduced. Overall, these results highlight a trade-off between recall and noise: while larger $K$ improves coverage, it may also harm performance due to irrelevant or redundant information. This underscores the importance of effective retrieval filtering and ranking strategies, rather than simply increasing the number of retrieved candidates.

\label{app:sec4.3}
\begin{table}[htbp]
\centering
\fontsize{8pt}{11pt}\selectfont
\setlength{\tabcolsep}{8pt}
\renewcommand{\arraystretch}{1.0}
\caption{Performance under different top-$K$ settings. The best result in each row is highlighted in bold.}
\label{tab:topk_results}
\begin{tabular}{lcccc}
\toprule
Method & top-5 & top-10 & top-15 & top-20 \\
\midrule
NaiveRAG & 0.4569 & 0.5023 & 0.5401 & \textbf{0.5449} \\
A-Mem~\cite{xu2026amem}    & 0.5757 & 0.6277 & \textbf{0.6428} & 0.6380 \\
MuRAG~\cite{chen-etal-2022-murag}    & 0.5527 & \textbf{0.5902} & 0.5665 & 0.5726 \\
NGM~\cite{fisher2025neural}      & 0.5049 & 0.6253 & \textbf{0.6277} & 0.6213 \\
\bottomrule
\end{tabular}
\end{table}

\section{Potential Risks}

The main risks of this work include potential misuse of the benchmark for surveillance applications and overgeneralization of results beyond English and synthetic data. The H2HMem dataset is released under a CC BY license to encourage broad use and reproducibility, but we urge researchers to deploy the benchmark only for benign applications such as meeting assistants and clinical documentation systems.

\end{document}